%% file: bare_jrnl_new_sample4.tex
\documentclass[a4paper,journal]{IEEEtran}
\usepackage{amsmath,amsfonts}
\usepackage{amsthm}
\usepackage{array}
\usepackage[caption=false,font=normalsize,labelfont=sf,textfont=sf]{subfig}
\usepackage{textcomp}
\usepackage{stfloats}
\usepackage{url}
\usepackage{verbatim}
\usepackage{graphicx}
\usepackage{cite}
\usepackage{dsfont}
\usepackage{bm} 
\usepackage{url}
\usepackage{balance}
\usepackage{mathtools}
\usepackage[dvipsnames]{xcolor} 
\usepackage{epstopdf}
\usepackage{graphicx}
\usepackage{subfig}
\usepackage{float}  
\usepackage{booktabs}
\usepackage{nicematrix}
\usepackage{makecell}
\usepackage{booktabs}

\usepackage{blindtext}
\usepackage{lipsum}

\usepackage[ruled,norelsize,linesnumbered]{algorithm2e}

\makeatletter
\newcommand{\removelatexerror}{\let\@latex@error\@gobble}
\makeatother

\SetKwComment{Comment}{/* }{ */}

\makeatletter
\renewcommand*\env@matrix[1][\arraystretch]{%
  \edef\arraystretch{#1}%
  \hskip -\arraycolsep
  \let\@ifnextchar\new@ifnextchar
  \array{*\c@MaxMatrixCols c}}
\makeatother

\usepackage{setspace}
\newcommand{\x}{\mathbf{x}}

\newcommand{\uu}{\mathbf{u}}
\newcommand{\y}{\mathbf{y}}
\newcommand{\vv}{\mathbf{v}}
\newcommand{\hh}{\mathbf{h}}
\newcommand{\ww}{\mathbf{w}}

\newcommand{\p}{\mathbf{p}}
\newcommand{\bP}{\mathbf{P}}
\newcommand{\ii}{\mathbf{i}}

\newcommand{\cF}{\mathcal{F}}
\newcommand{\bF}{\mathbf{F}}
\newcommand{\g}{\mathbf{g}}
\newcommand{\A}{\mathbf{A}}
\newcommand{\B}{\mathbf{B}}

\newcommand{\K}{\mathbf{K}}
\newcommand{\M}{\mathbf{M}}
\newcommand{\LL}{\mathbf{L}}
\newcommand{\cL}{\mathcal{L}}

\newcommand{\R}{\mathbb{R}}
\newcommand{\Q}{\mathbf{Q}}
\newcommand{\bH}{\mathbf{H}}
\newcommand{\bR}{\mathbf{R}}

\newcommand{\NU}{\bm{\nu}}
\newcommand{\uM}{\underline{\mathcal{M}}}
\newcommand{\sht}{s_{\mathrm{ht}}}

\newcommand{\hbx}{\hat{\mathbf{x}}}
\newcommand{\CH}{\mathcal{C}\mathcal{H}}
\newcommand{\IP}{\mathcal{I}\mathcal{P}}
\newcommand{\OB}{\mathcal{O}\mathcal{B}}
\newcommand{\lsdeng}{\hspace{-0.1cm}=\hspace{-0.1cm}}
\newcommand{\lsjia}{\hspace{-0.08cm}+\hspace{-0.08cm}}
\newcommand{\lsjian}{\hspace{-0.08cm}-\hspace{-0.08cm}}
\newcommand{\avunit}{[\mathrm{rad}\cdot\mathrm{s}^{-1}]}
\newcommand{\vunit}{[\mathrm{m}\cdot\mathrm{s}^{-1}]}
\newcommand{\munit}{[\mathrm{m}]}
\newcommand{\tunit}{[\mathrm{s}]}
\newcommand{\funit}{[\mathrm{s}^{-1}]}
\newcommand{\aunit}{[\mathrm{rad}]}

\newtheorem{theorem}{Theorem}
\newtheorem{definition}{Definition}

\newtheorem{remark}{Remark}

\newcommand{\hlt}[1]{{\color{BrickRed} #1}}

\newcommand{\chang}[1]{{\color{black} #1}}
\newcommand{\sven}[1]{{\color{black} #1}}
\newcommand{\guido}[1]{{\color{black} #1}}

\DeclareMathOperator*{\diag}{diag}
\DeclareMathOperator*{\vect}{vec}

\allowdisplaybreaks
\graphicspath{{./figure/}}

\begin{document}

\title{\guido{Cooperative Relative Localization in MAV Swarms with Ultra-wideband Ranging}}

\author{Changrui Liu\textsuperscript{$\text{*}$}, Sven Pfeiffer\textsuperscript{$\dagger$} and Guido C.H.E. de Croon\textsuperscript{$\dagger$} 
\thanks{\textsuperscript{$\text{*}$} C. Liu is with the Delft Center for Systems and Control, Delft University of Technology, and was with the Micro Air Vehicle Laboratory, Faculty of Aerospace Engineering, Delft University of Technology, Delft, The Netherlands {\tt\small C.Liu-14@tudelft.edu},}
\thanks{\textsuperscript{$\dagger$} The authors are with the Micro Air Vehicle Laboratory, Faculty of Aerospace Engineering, Delft University of Technology, 2628CD Delft, The Netherlands {\tt\small \{s.u.pfeiffer,g.c.h.e.decroon\}@tudelft.nl}}}

\markboth{Journal of \LaTeX\ Class Files,~Vol.~XX, No.~X, X~2022}%
{Liu \MakeLowercase{\textit{et al.}}: \guido{Cooperative Relative Localization in MAV Swarms with Ultra-wideband Ranging}}

\IEEEpubid{0000--0000/00\$00.00~\copyright~2022 IEEE}

\maketitle

\thispagestyle{empty} 

\input{content_file/0-abstract.tex}
\input{content_file/1-introduction.tex}

\input{content_file/2-Preliminaries_Problem_Formulation.tex}
\input{content_file/3-Cooperative_Relative_Localization.tex}

\input{content_file/4-main_approaches.tex}

\input{content_file/5-simulation_results.tex}

\input{content_file/6-conclusion.tex}

{


}


\input{reference.tex}

\vfill

\end{document}

%% file: content_file/0-abstract.tex
\begin{abstract}
Relative localization (RL) is essential for the successful operation of micro air vehicle (MAV) swarms. Achieving accurate 3-D RL in infrastructure-free and GPS-denied environments with only distance information is a challenging problem that has not been satisfactorily solved. In this work, based on the range-based peer-to-peer RL using the ultra-wideband (UWB) ranging technique, we develop a novel UWB-based cooperative relative localization (CRL) solution which integrates the relative motion dynamics of each host-neighbor pair to build a unified dynamic model and takes the distances between the neighbors as \textit{bonus information}. Observability analysis using differential geometry shows that the proposed CRL scheme can expand the observable subspace compared to other alternatives using only direct distances between the host agent and its neighbors. In addition, we apply the kernel-induced extended Kalman filter (EKF) to the CRL state estimation problem with the novel-designed Logarithmic-Versoria (LV) kernel to tackle heavy-tailed UWB noise. Sufficient conditions for the convergence of the fixed-point iteration involved in the estimation algorithm are also derived. Comparative Monte Carlo simulations demonstrate that the proposed CRL scheme combined with the LV-kernel EKF significantly improves the estimation accuracy owing to its robustness against both the measurement outliers and incorrect measurement covariance matrix initialization. Moreover, with the LV kernel, the estimation is still satisfactory when performing the fixed-point iteration only once for reduced computational complexity.
\end{abstract}


%% file: content_file/1-introduction.tex
\section{Introduction}
\IEEEPARstart{M}{icro} Air Vehicles (MAVs), with higher agility and a lighter design, have been widely used in many real-world operations such as surveillance \cite{scaramuzza2014vision}, exploration \cite{bahr2009consistent}, and urban construction \cite{augugliaro2014flight}. Swarm behaviors observed in the natural world (e.g. starlings and bees) motivate the development of \guido{MAV swarms} \sven{to compensate} the limited power and mobility of a single MAV agent. \sven{Swarming allows robots to perform more complex tasks} (e.g., cooperative transportation \cite{menouar2017uav} and gas detection \cite{duisterhof2021sniffy}) where efficiency and \chang{system redundancy} are of major consideration.
\begin{figure}[t]
    \centering
    \includegraphics[width=0.9\linewidth]{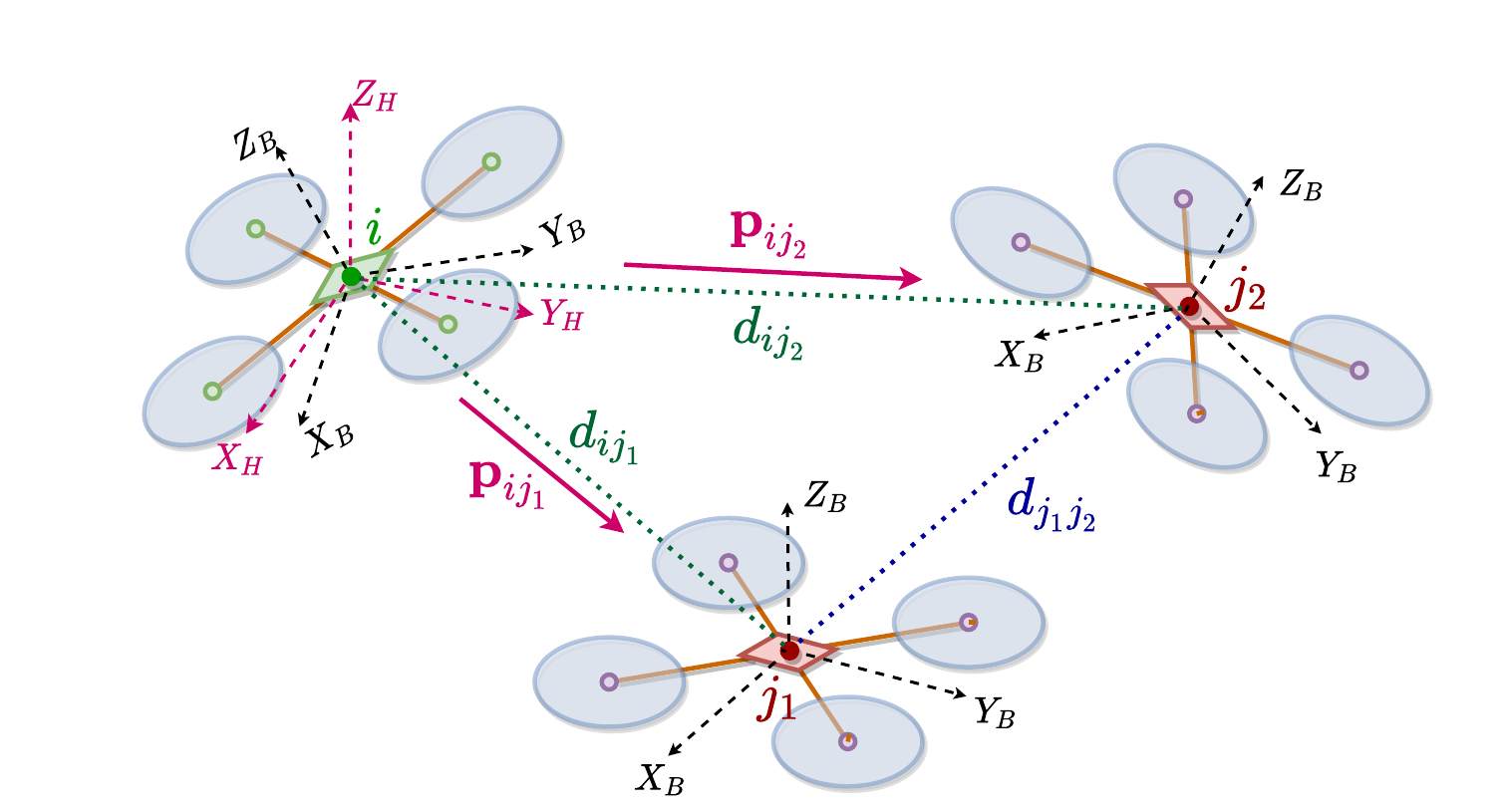}
    \caption{Cooperative relative localization problem, where agent $i$ aims to localize its neighbors $j_1$ and $j_2$ in its body centered horizontal frame, i.e., agent $i$ needs to compute $\p_{ij_1}$ and $\p_{ij_2}$.}
    \label{fig:mavproblem}
\end{figure}

Successful operation of \sven{an} MAV-swarm requires an accurate relative localization (RL) scheme which then provides essential position feedback among agents for collision avoidance \cite{coppola2018board} and formation control \cite{guo2019ultra}, both of which are prerequisites for higher-level objectives. RL schemes can be broadly categorized as indirect and direct ones. Indirect solutions compute relative position using global positions in a shared frame, typical examples use GPS \cite{dong2016time, kang2016distance}, optical motion tracking systems \cite{dudzik2020application} and anchor-dependent positioning systems with fixed beacons or ground stations \cite{fang2020graph, ruan2021cooperative, nguyen2016ultra}.
\sven{\guido{The performance} of GPS degrades substantially in indoor and urban environments, and localization relying on external devices requires additional investments (time and money) for setting up the area of operation.}
Consequently, it is favorable to develop \textit{GPS-denied} and \textit{infrastructure-free} RL solutions that provide position estimation \textit{directly} through inter-agent sensing and communication. 

\guido{For} direct RL schemes where onboard sensors play a more dominant role, the visual-odometry methods \cite{chudoba2014localization, jurevivcius2019robust} which extract relative positions using cameras are the most straightforward. \guido{However, the high computational cost and limited field-of-view of the vision-based methods are not ideal for MAVs.} Excluding visual information, fusing IMU with other ranging sensors is effective and implementable. Ultra-wideband (UWB) technology has recently drawn great attention in aerial robots localization tasks \cite{guo2019ultra, guler2020peer, li2020autonomous} owing to its \sven{superior} communication capability and ranging accuracy compared to infrared sensors, microphones, and Bluetooth \cite{lee2007comparative}. Therefore, equipping MAVs with UWB tags is a lightweight and economic solution for the RL purpose in swarming. In terms of UWB-based RL, the underlying problem is 3-D position estimation with inter-agent \textit{distance} measurements. Some previous attempts either consider 2-D scenarios by using altimeters \cite{li2020autonomous, guo2019ultra} or use multiple UWB tags to provide bearing information \cite{guler2020peer, shalaby2021relative}. The \textit{purely distance-based} 3-D RL problem with only one UWB tag equipped on each MAV has not been satisfactorily solved, and typical solutions relying on graph rigidity theory \cite{liu2020distributed} or optimization \cite{trawny20073d, jiang20193} not only fail to address noise issues in the measurements but also rely on a common reference frame \guido{which is not available to MAVs flying in unknown environments}. In fact, due to multi-pathing and non-line-of-sight (NLoS) effects, the UWB distance measurements are noisy, thus making filtering-based estimation algorithms more suitable for accurate localization. Considering the limited onboard computational resources and the nonlinear nature of the relative motion dynamics, the estimation algorithm design for the distance-based 3-D RL becomes more challenging.

In this work, we aim to address the RL problem for MAV swarms by determining for each agent in the swarm the relative position of its neighboring agents in its body-centered horizontal frame.
Though the basic peer-to-peer RL (i.e., performing position estimation for each of the neighbors individually \sven{as done by Li et al. \cite{li2020autonomous}}) suffices to solve this problem, the benefits of swarming are not fully exploited. The main contributions of this work are twofold. First, we propose a novel cooperative relative localization (CRL) scheme which integrates the relative motion dynamics of each of the neighbors to build a unified dynamic model and also takes the distances between the neighbors as \textit{bonus information}. The resulting CRL can account for the correlation induced by the same velocity input as well as improve localization accuracy in general by introducing more measurements. Second, inspired by the kernel-based filtering methods \cite{chen2017maximum}, we propose a maximum Logarithmic-Versoria criterion-based extended Kalman filtering (MLVC-EKF) for the \chang{state estimation in localization tasks} to handle the heavy-tailed noise in UWB distance measurements. The proposed filtering strategy substantially improves robustness against measurement outliers, especially for the CRL framework where more measurements are present.

The rest of the paper is organized as follows. In Section \ref{sec:2-pre}, we present mathematical convention and some theoretical backgrounds. Then, Section \ref{sec:3-prob} formulates the state estimation problem for the CRL. In the following Section \ref{sec:4-mainapp}, we provide the MLVC-EKF algorithm and analyze the CRL problem in detail. The simulation results and the discussions are given in Section \ref{sec:5-simulation}. Lastly, we conclude the work and delineate future research directions in \ref{sec:6-conclusion}.

%% file: content_file/2-Preliminaries_Problem_Formulation.tex
\section{\chang{Preliminaries \& Problem Formulation}}
\label{sec:2-pre}
\chang{In this section, we will first present some mathematical conventions used throughout the paper. Subsequently, we will present the details of the relative motion dynamics which will be used for further CRL modeling. Finally, we elaborate on the UWB noise modeling where we consider various practical issues (e.g., NLoS, multi-pathing, and transmission delay).}

\input{content_file/2.1-Notation.tex}
\input{content_file/2.2ex-p2p-dynamics.tex}
\input{content_file/2.3-UWBnoise.tex}

%% file: content_file/2.1-Notation.tex
\subsection{Notation and Definitions}
\guido{Let $\x \in \R^n$ denote a vector with $\x[k]$ and $\| {\x} \|_p \; (p \in \{1,2,\infty\})$ being its $k$-th element and vector $p$-norm. $\M\in\R^{n\times m}$ is a matrix with $\M^\top$, $\M[k]$ and $\M^\top[k]$ being its transpose, $k$-th row and $k$-th column, respectively. Similarly, $\| {\M} \|_p \; (p \in \{1,2,\infty\})$ denotes the induced matrix $p$-norm. In the sequel, all norms will be referred to as $2$-norm if not explicitly specified. The set of positive integers up to $n$ 
is $\mathbb{Z}_n^{+}$, and $\mathbb{Z}_{j}^{i}$ denotes non-negative integers from $i$ to $j \; (0\leq i\leq j)$. 

The minimum (maximum) eigenvalue of a matrix $\M$ is denoted as $\lambda_{\text{min}}(\M)$ ($\lambda_{\text{max}}(\M)$). Given $\x, \y \in \R^n$, $\x/\y$ means element-wise division.} \sven{The symbol $\otimes$ denotes the Kronecker product.} $\mathbf{0}_n$, $\mathds{1}_n$, and $\mathds{I}_n$ are $n$-dimensional zero vector, one vector, and identity matrix, respectively. The $n$-dimensional basis row vector with the $i$-th entry being 1 while all other entries being 0 is denoted as $\mathbf{b}^n_i$. Further, we define operators $\diag$ and $\vect$ for matrices (vectors) as: 
\begin{align*}
    \begin{aligned}
    \small
	&\hspace{-0.1cm} \textstyle\diag \{ M_i, M_j\} = \begin{bmatrix} M_i & \mathbf{0} \\  \mathbf{0} & M_j \end{bmatrix},
    &\hspace{-0.2cm} \textstyle\vect \{ M_i, M_j\} = \begin{bmatrix} M_i  \\ M_j \end{bmatrix}, 
        \end{aligned}
\end{align*}
\begin{align*}
    \begin{aligned}
    \small
    &\hspace{-0.1cm} \textstyle\diag_{k=i}^j \{ M_k\} = \begin{bmatrix} M_i & &\\[-0.1cm] & \ddots & \\[-0.1cm] & & M_j \end{bmatrix}, \;
    &\hspace{-0.2cm} \textstyle\vect_{k=i}^j \{ M_k\} = \begin{bmatrix} M_i \\[-0.1cm] \vdots \\[-0.1cm] M_j \end{bmatrix}\guido{.}
    \end{aligned}
\end{align*}

For convenience, the trigonometric functions $\sin(\cdot)$, $\cos(\cdot)$ and $\tan(\cdot)$ are abbreviated as $s(\cdot)$, $c(\cdot)$ and $t(\cdot)$, respectively. The Cholesky decomposition for matrices is denoted as a function $\mathcal{C}\mathcal{H}(\cdot)$. Besides, the probability \chang{distribution} function (PDF) of a random variable $X$ obeying Gaussian distribution is denoted as \guido{$G(x; \mu,\sigma)$, with $\mu$} being the mean and $\sigma$ being the standard deviation. For Gamma distribution, \guido{we use $\gamma(x; k,\varepsilon) = [\Gamma(k)]^{-1}\varepsilon^k x^{k-1}\mathrm{exp}(-\varepsilon x)$}, where $\Gamma(\cdot)$ is the gamma function. Lastly, the uniform distribution on interval $[a, b]$ is denoted as $U(a,b)$.

%% file: content_file/2.2ex-p2p-dynamics.tex
\subsection{Peer-to-peer Relative Motion Dynamics}
\chang{We first introduce the relative motion dynamics for a pair of agents, which is the basic building block of the CRL models in later stages.} For a given MAV, the 3 axes of the body-fixed body frame ($\cF_{B}$) originate from the center of gravity of the robot with the $X$-axis pointing forward, $Z$-axis \chang{being} aligned to the direction of the thrust, and $Y$-axis completing the frame according to the right-handed convention. In addition, to simplify the localization problem, we mainly deal with the body-centered \textit{horizontal frame} ($\cF_{H}$) with its $Z$-axis being always \textit{perpendicular} to the ground plane \sven{\cite{li2020autonomous}}. With $\cF_{B}$ and $\cF_{H}$ defined, the orientation of both frames, as well as the transformation between the frames, can be determined using Euler angles (i.e., $\phi$, $\theta$, $\psi$) and the corresponding rotation matrices $\bR_X(\phi)$, $\bR_Y(\theta)$, and $\bR_Z(\psi)$.

In this work, the relative position is resolved in the horizontal frame $\cF_{H}$, and all related variables should be properly transformed accordingly. First, the velocity $\vv$ is measured in $\cF_{B}$ and it needs to be expressed in $\cF_{H}$. The rotation matrix transforming any vector in $\cF_{B}$ to $\cF_{H}$ is given as
\begin{equation}
\small
    \label{eq:rotation_h2b}
    \hspace{-0.2cm}
    ^{H}\bR_{B} = \bR_Y(\theta)\bR_X(\phi) = 
    \begin{bmatrix}
    c(\theta) & s(\theta)s(\phi) & s(\theta)c(\phi) \\
    0 & c(\phi) & -s(\phi) \\
    -s(\theta) & c(\theta)s(\phi) & c(\theta)c(\phi)
    \end{bmatrix}.
\end{equation}
Second, the gyroscope measures the instantaneous angular velocity $\mathbf{\Omega}$ in $\cF_{B}$, to which the Euler angle rate should be properly related, and the formula is given as
\begin{equation}
\small
    \label{eq:rotation_angular}
    \begin{bmatrix}
        \dot{\phi} \\
        \dot{\theta} \\
        \dot{\psi}
    \end{bmatrix}
    =
    \begin{bmatrix}
        1 & s(\phi)t(\theta) & c(\phi)t(\theta) \\
        0 & c(\phi) & -s(\phi) \\
        0 & s(\phi)/c(\theta) & c(\phi)/c(\theta)
    \end{bmatrix}
    \begin{bmatrix}
        {\Omega}_x \\
        {\Omega}_y \\
        {\Omega}_z
    \end{bmatrix}.
\end{equation}
In particular, we are interested in computing the heading rate $\dot{\psi}$, to which the changing of the horizontal frame is related. Note that the conversion in \eqref{eq:rotation_h2b} and \eqref{eq:rotation_angular} requires the roll ($\phi$) and pitch ($\theta$), both of which can be accurately estimated by IMU.

For any agent pair $(i,j)$, the peer-to-peer RL problem is defined as estimating the relative position of agent $j$ with respect to agent $i$ in agent $i$'s horizontal frame $\cF^i_{H}$. The state vector is thus given as $\x_{ij} = [\psi_{ij}, \p^\top_{ij}]^\top$, where $\psi_{ij}$ is the relative heading and $\p_{ij} = [x_{ij}, y_{ij}, z_{ij}]^\top$ is the relative position in $\cF^i_{H}$. The control input of any agent $i$ is defined as $\uu_i = [\dot{\psi}_i, \vv^\top_i]^\top$ where $\vv_i = [v_{x,i}, v_{y,i}, v_{z,i}]^\top$ is the velocity input \guido{in the body frame $\cF^i_{H}$ of agent $i$}, and the control input for agent pair $(i,j)$ is $\uu_{ij} = \vect\{\uu_i, \uu_j\}$. Then, the relative motion dynamics for agent pair $(i,j)$ can be derived using Newton's law as
\begin{equation}
    \label{eq:dy_pair_compact}
    \dot{\x}_{ij} = \g(\x_{ij}, \uu_{ij}) = 
    \begin{bmatrix}
        \dot{\psi}_j - \dot{\psi}_i \\
        \bR(\psi_{ij})\vv_j - \vv_i - \dot{\psi}_i S\p_{ij}
    \end{bmatrix},
\end{equation}
where the matrices $\bR(\psi_{ij})$ and $S$ are given as
\begin{equation*}
\small{
    \bR(\psi_{ij}) = 
    \begin{bmatrix}
        c(\psi_{ij}) & -s(\psi_{ij}) & 0 \\
        s(\psi_{ij}) & c(\psi_{ij}) & 0 \\
        0 & 0 & 1 
    \end{bmatrix},
    \;
    S = 
    \begin{bmatrix}
        0 & -1 & 0 \\
        1 & 0 & 0 \\
        0 & 0 & 0
    \end{bmatrix}.
    }
\end{equation*}

%% file: content_file/2.3-UWBnoise.tex
\subsection{Heavy-tailed UWB Measurement Noise \& Delay Effects}
Ideally, the true distance $d^\ast_{ij}$ between agent pair $(i,j)$ is given in terms of its corresponding relative position $\p^\ast_{ij}$ as
\begin{equation}
    \label{eq:dis_model_ideal}
    d^\ast_{ij} = \| \p^\ast_{ij} \|.
\end{equation}
However, the measured distance deviates from the true value due to many practical factors. First, the distance measuring scheme in UWB is two-way ranging (TWR), whose noise model can be ideally assumed to be Gaussian considering only line-of-sight (LoS) scenarios. A more realistic noise model accounting for both multi-pathing and NLoS effects leads to a heavy-tailed distribution which is a linear combination of a Gaussian distribution and a Gamma distribution and other additional terms for outliers. We adopt the PDF of the simplified model \cite{pfeiffer2021computationally} for the noise $\nu$ as
\begin{equation}
    \label{eq:noise_pdf_sven}
    f(\nu) = \frac{1}{1+\sht}G(\nu; \sht \mu, \sigma) + \frac{\sht}{1+\sht}\gamma(\nu; k, \lambda),
\end{equation}
where \sven{$\sht \geq 0$} is a scale factor that \chang{is} chosen based on \chang{the} real measurement data.

\begin{figure}
    \centering
    \includegraphics[width=0.5\linewidth]{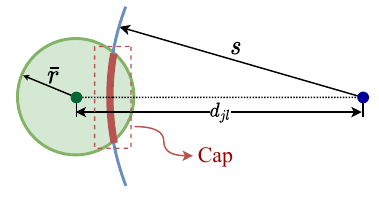}
    \caption{2-D visualization of the spherical cap which is marked in red.}
    \label{fig:cap}
\end{figure}

\begin{figure}[h]
    \centering
    \includegraphics[width=0.85\linewidth]{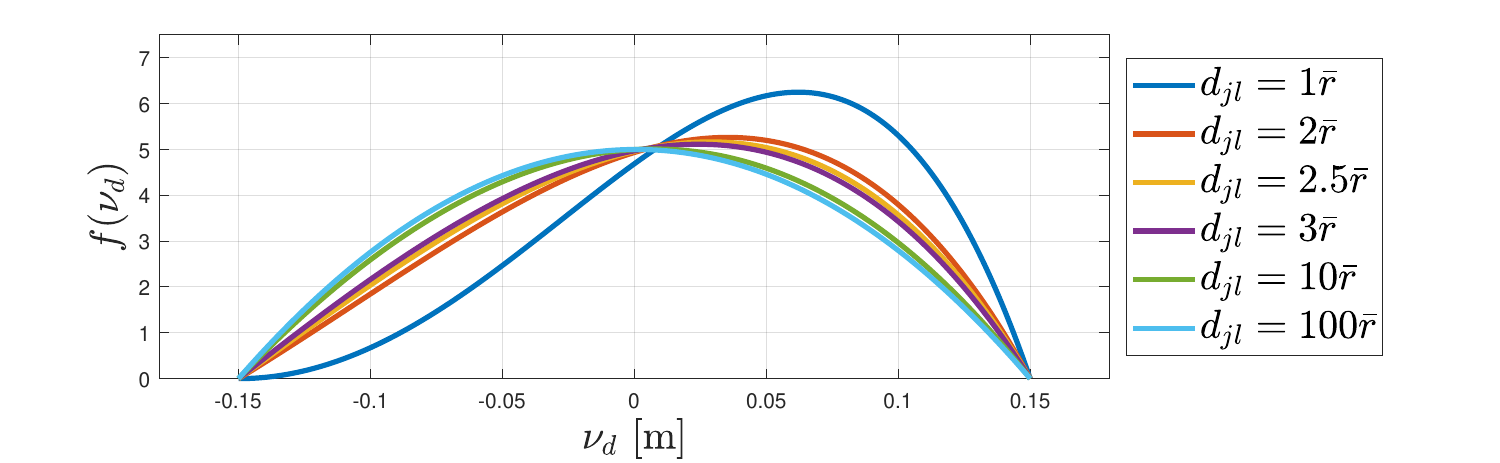}
    \caption{PDFs of noise $\nu_d$}
    \label{fig:delaynoise}
\end{figure}

Another source of measurement error comes from communication delay, especially for the indirect distance measurements between neighbors. Due to the limited bandwidth, the distances cannot be measured and transmitted simultaneously. Therefore, for agent $i$ which can only get the distance between its neighbors $j$ and $l$ through the communication between agent $j$ or $l$, the instantaneous true distance $\| \p^\ast_{jl} \|$ at the transmitting time $t_{\mathrm{tr}}$ differs from the transmitted $\| \p_{jl} \|$ measured at time $t_{\mathrm{ms}}$ due to the \chang{unknown} motion of agent $j$ and $l$ during the time interval $[t_{\mathrm{ms}}, t_{\mathrm{tr}}]$. Assume the length of the \guido{time interval} is upper bounded by $\bar{\eta}$, and the relative speed between any pair of agents is also bounded by $\bar{v}$, we have $\p^\ast_{jl} = \p_{jl} + \p_e = \p_{jl} + (\alpha\bar{\eta}\bar{v})\ii$ with $\alpha \in [0,1]$ and $\ii$ a \textit{unit} vector. For modeling simplicity, we further assume that $\ii$ has an equal probability of pointing in any direction and $\alpha$ is \textit{uniformly} distributed on $[0,1]$. Then, $\p_e$ has equal probability pointing at any point in a sphere \chang{of} radius $\bar{r} = \bar{\eta}\bar{v}$. As a result, the probability of having a distance error $\nu_d = s - d_{jl}$ is proportional to the area of the spherical cap of a sphere \chang{of} radius $s \in [d_{jl} - \bar{r}, d_{jl} + \bar{r}]$ intersected by a sphere with radius $\bar{r}$, \guido{where $d_{jl} = \|\p_{jl}\|$ is the distance between agent $j$ and $l$ with measurement delay. The area of the spherical cap can be \textit{approximated} as
\begin{equation}
    \label{eq:spherical_cap}
    a(\nu_d; d_{jl}, \bar{r}) = \pi[-(\nu_d^2 + 2\nu_dd_{jl}-\bar{r}^2)^2 + 4d_{jl}^2\bar{r}^2 ],
\end{equation}
and the 2-D visualization of the cap is given in Fig. \ref{fig:cap}.}
Following \eqref{eq:spherical_cap}, the normalized PDF of $\nu_d$ is thus given as
\chang{
\begin{equation}
    \label{eq:noise_pdf_changrui}
    \hspace{-0.35cm}
    f(\nu_d) = \frac{15}{16 \pi \bar{r}^3(5d^2_{jl} - \bar{r}^2)} a(\nu_d; d_{jl}, \bar{r}),\; (\nu_d \in [-\bar{r}, \bar{r}]).
\end{equation}
}
\sven{The severity of the delay effects hinges on the velocity of the MAV and the density of the swarm. On the one hand, \guido{if the MAVs fly at high speed such that the relative velocity can be high enough}, then $\bar{v}$ can be large. On the other hand, if the swarm is too dense such that the number of neighbors of each agent is big, then $\bar{\eta}$ can be quite large due to limited bandwidth. In Fig. \ref{fig:delaynoise}, we present some representative PDFs when $\bar{r} = 0.15\munit$.}
\begin{remark}
    The noise $\nu$ and $\nu_d$ are independent, which explains why they are considered separately instead of being combined into a single type of noise with a unified PDF.
\end{remark}
\begin{remark}
     The approximation in \eqref{eq:spherical_cap} is valid under the condition that $\bar{r} \ll d_{jl}$, \chang{which can be relaxed since that the noise distribution does not differ too much when $\bar{r} \leq 3d_{jl}$.} \guido{In most cases, the condition $\bar{r} > d_{jl}$ holds given that the number of neighbors for any agent is limited such that $\bar{\eta}$ does not permit a large value and two MAVs do not fly in a dangerous mode in which the inter-agent distance is too short such that the random motion within a short time interval is likely to cause a collision (i.e., $d_{jl} \leq \bar{r}$).}
\end{remark}

%% file: content_file/3-Cooperative_Relative_Localization.tex
\section{\chang{Cooperative Relative Localization}}
\label{sec:3-prob}
\chang{This section presents 3 different CRL models which serve as the main problem of interest in this work. Next, to shed light on the state estimation, we present the observability analysis for the built models using some elementary differential geometric tools.}

\input{content_file/3.1ex-ThreeCRL-Models.tex}
\input{content_file/3.2ex-Differential_Geometry.tex}

\input{content_file/3.3ex-Observability.tex}

%% file: content_file/3.1ex-ThreeCRL-Models.tex
\subsection{Modeling of CRL Schemes} 
Having the peer-to-peer relative motion dynamics established as in \eqref{eq:dy_pair_compact}, we \chang{are able to delve into} the CRL modeling. Consider a swarm of $N$ agent, and for agent $i$, the set of its neighbors is $\mathcal{N}_{i}$ with $|\mathcal{N}_{i}| = N_{i}$. We explicitly express $\mathcal{N}_{i}$ as $\{j_{\alpha} \in \mathbb{Z}^+_{N}|j_{\alpha} \neq i, j_{\alpha} < j_{\alpha+1}\;(\alpha = 1,2,\cdots,N_{i})\}$.
To build an integrated dynamic model, we augment the state $\x_{ij_\alpha}$ of agent pair $(i,j_\alpha)$, and the augmented state $\bar{\x}_i$ is given as
\begin{equation}
    \label{eq:aug_state}
    \bar{\x}_i = \textstyle\vect^{N_i}_{\alpha=1}\{\x_{ij_\alpha}\}.
\end{equation}
We also augment the control input $\uu_i$ to form $\bar{\uu}$ as
\begin{equation}
    \label{eq:aug_control}
    \bar{\uu}_i = \vect\{\uu_i, \textstyle\vect^{N_i}_{\alpha=1} \{\uu_{j_\alpha}\} \}.
\end{equation}
Following \eqref{eq:dy_pair_compact}, we can define $\bar{\g}^c_i(\bar{\x}_i, \bar{\uu})$ similarly as
\begin{equation}
    \label{eq:aug_control_matrix_compact}
    \bar{\g}_i(\bar{\x}_i, \bar{\uu}_i) = \textstyle\vect^{N_i}_{\alpha=1}\{\g(\x_{ij_\alpha}, \uu_{ij_\alpha})\} ,
\end{equation}
and the augmented system directly follows as
\begin{equation}
    \label{eq:dy_aug_compact}
    \dot{\bar{\x}}_i = \bar{\g}_i(\bar{\x}_i, \bar{\uu}_i).
\end{equation}

For the measurement model of CRL, we first augment the direct distance model between agent $i$ and each of its neighbors
\begin{equation}
    \label{eq:dis_direct}
    \hh^\mathrm{d}_i(\bar{\x}_i) = \textstyle\vect^{N_i}_{\alpha=1}\{\|\p_{ij_\alpha}\|\}.
\end{equation}
Then, the augmented direct measurement considering only the noise $\nu$ is given as
\begin{equation}
    \label{eq:dis_direct_noisy}
    \y^\mathrm{d}_i = \hh^\mathrm{d}_i(\bar{\x}_i) + \bm{\nu},
\end{equation}
where \guido{$\bm{\nu} = \textstyle\vect^{N_i}_{\alpha=1}\{\nu_\alpha\}$} is the augmented noise. On the other hand, the augmented indirect distances transmitted by agent $j_\alpha$ are given as
\begin{equation}
    \label{eq:dis_indirect_neighborwise}
    \hh^\mathrm{id}_i(\bar{\x}_i; j_\alpha) = \textstyle\vect_{\{l \in \mathcal{N}_{i} \cap \mathcal{N}_{j_\alpha} \}}\{ \|\p_{ij_\alpha} - \p_{il}\| \},
\end{equation}
and the augmented indirect distances considering all its neighbors follows as
\begin{equation}
    \label{eq:dis_indirect_all}
    \hh^\mathrm{id}_i(\bar{\x}_i) = \textstyle\vect^{N_i}_{\alpha = 1}\{ \hh^\mathrm{id}_i(\bar{\x}_i; j_\alpha)\}.
\end{equation}
The final noisy indirect measurement model is given as
\begin{equation}
    \label{eq:dis_indirect_noisy}
    \y^\mathrm{id}_i = \hh^\mathrm{id}_i(\bar{\x}_i) + \bm{\nu}_a
\end{equation}
where \chang{$\bm{\nu}_a = \textstyle\vect^{p_{\mathrm{id}}}_{\beta=1}\{\nu_\beta + \nu_{d,\beta}\}$} with $p_{\mathrm{id}} = \sum^{N_i}_{\alpha=1}|\mathcal{N}_{i} \cap \mathcal{N}_{j_\alpha}|$.

Moreover, we assume that all velocity inputs are imperfect due to actuator noise $\Delta \uu_i$ for agent $i$, with which we also define $\Delta \uu_{ij_\alpha} = \vect\{\Delta \uu_i, \Delta \uu_{j_\alpha}\}$ and $\Delta \bar{\uu}_i = \vect\{\Delta \uu_i, \textstyle\vect^{N_i}_{\alpha=1} \{\Delta \uu_{j_\alpha}\} \}$. Referring to \eqref{eq:dy_aug_compact}, \eqref{eq:dis_direct_noisy} and \eqref{eq:dis_indirect_noisy}, the full-CRL (fCRL) model with the augmented dynamics and both direct and indirect measurements is given as
\begin{equation}
\label{eq:fcrl}
    \text{fCRL}:\left \{
    \begin{aligned}
    \dot{\bar{\x}}_i &= \sven{\bar{\g}_i(\bar{\x}_i, \bar{\uu}_i + \Delta\bar{\uu}_i)} \\
    \begin{bmatrix}
        \y^\mathrm{d}_i \\
        \y^\mathrm{id}_i
    \end{bmatrix}
    &=
    \begin{bmatrix}
        \hh^\mathrm{d}_i(\bar{\x}_i) \\
        \hh^\mathrm{id}_i(\bar{\x}_i)
    \end{bmatrix}
    +
    \begin{bmatrix}
        \bm{\nu} \\
        \bm{\nu}_a
    \end{bmatrix}
    \end{aligned}
    \right. .
\end{equation}

For comparative study, we also introduce the following two models: i) the half-CRL (hCRL) model with the augmented dynamics and only direct measurements, ii) the non-CRL (nCRL) model which is simply \textit{a collection} of the peer-to-peer dynamics in \eqref{eq:dy_pair_compact} for agent pairs $(i,j_\alpha)$ without state and input augmentation, \sven{$\forall j_\alpha \in \mathcal{N}_i$}. The details are given below:
\begin{equation}
\label{eq:hcrl}
    \hspace{-3.3cm}
    \text{hCRL}:\left \{
    \begin{aligned}
    \dot{\bar{\x}}_i &= \sven{\bar{\g}_i(\bar{\x}_i, \bar{\uu}_i + \Delta\bar{\uu}_i)} \\
        \y^\mathrm{d}_i
    &=
        \hh^\mathrm{d}_i(\bar{\x}_i)
    +
        \bm{\nu} 
    \end{aligned}
    \right. ,
\end{equation}
\begin{equation}
\label{eq:ncrl}
    \text{nCRL}:\bigg\{ \left \{
    \begin{aligned}
    \dot{\x}_{ij_\alpha} &= \sven{\g(\x_{ij_\alpha}, \uu_{ij_\alpha} + \Delta \uu_{ij_\alpha})}\\
    y_{ij_\alpha} &= \|\p_{ij_\alpha}\| + \nu_\alpha
    \end{aligned}
    \right. \bigg\}_{\{j_\alpha \in \mathcal{N}_i\}}.
\end{equation}
\sven{
\begin{remark}
    The 3 dynamic models (i.e., fCRL, hCRL, and nCRL) are all continuous-time nonlinear control systems. In addition, the built models are \textit{stochastic} in nature due to actuator and measurement noise.
\end{remark}
}
\guido{
\begin{remark}
    The nCRL model is equivalent to the peer-to-peer relative localization as investigated in \cite{li2020autonomous, coppola2018board}. 
\end{remark}
}

%% file: content_file/3.2ex-Differential_Geometry.tex
\subsection{Knowledge from Differential Geometry}
\guido{We present some basic geometric concepts following \cite{isidori1985nonlinear}.} Consider a time-varying (TV) nonlinear control system
\begin{equation}
\label{eq:sys_math}
\left \{
\begin{aligned}
    \dot{\x} &= \g(\x, \uu, t) \\
    \y &= [h_1(\x), h_2(\x), \cdots, h_p(\x)]^\top
\end{aligned}
\right. ,
\end{equation}
where $\x$ evolves in a $\mathcal{C}^\infty-\text{manifold}$ $\mathcal{M}$ of dimension $n$, \chang{$\uu \in \mathcal{U} \subseteq \mathbb{R}^m$ is the control input}, $\g(\x, \uu, t)$ \guido{is} the TV control-coupled vector \guido{field} in $\mathcal{M}$, and $h_i(\x)$ are time-invariant (TI) scalar fields also defined on $\mathcal{M}$. Besides, all functions are $\mathcal{C}^\infty$ \guido{(i.e., smooth/infinitely differentiable)} functions of their arguments. The time $t$ belongs to an open interval $\mathcal{I}\subseteq \mathbb{R}$, and we denote \chang{$\uM = \mathcal{I}\times \mathcal{M}\times \mathcal{U}$}. To analyze the above system, we present some basic geometric tools.
\begin{definition}[Lie Derivative]
Given a scalar function $h(\x):\mathcal{M} \mapsto \mathbb{R}$, and a vector field $\g(\x,\uu, t): \uM \mapsto \mathcal{M}$. The \textit{Lie derivative} of \chang{the} covector $\mathrm{d}h$ along the vector field $\g(\x,\uu, t)$ is defined as
\begin{equation}
    \label{eq:lie_derivative}
    \cL_{\g}\mathrm{d}h = \mathrm{d}(\cL_{\g}h) = \mathrm{d}(\langle \frac{\partial h}{\partial \x}, \g(\x,\uu,t)\rangle),
\end{equation}
where $\mathrm{d}h$, being the dual of $\g(\x,\uu,t)$, is a special type of covector called \textit{exact differential}, \sven{and $\langle \cdot,\cdot\rangle$ denotes the inner product of two vector fields}. Further, a set of \textit{covectors} $\mathrm{d} h_i$ can form a \textit{codistribution} $\Xi = \text{span}\{\mathrm{d} h_i\}$, and the Lie derivative of the codistribution is $\cL_{\g}\Xi = \text{span}\{\cL_{\g}\xi | \forall \xi \in \Xi\}$. \guido{Taking the Lie derivative on a codistribution is useful when recursively computing the observability codistribution, especially for systems with multiple output functions.}
\end{definition}

To analyze the time-varying system in \eqref{eq:sys_math}, we introduce the \textit{augmented Lie derivative} \cite{martinelli2022extension} for a covector $\mathrm{d}h$ as
\begin{equation}
    \label{eq:lie_newoperator}
    \dot{\cL}_{\g}\mathrm{d}h = \cL_{\g}\mathrm{d}h + \frac{\partial \mathrm{d}h}{\partial t}.
\end{equation}
Besides, this new derivative permits a recursive operation as
\begin{equation}
    \dot{\cL}^r_{\g}\mathrm{d}h = \dot{\cL}_{\g}(\dot{\cL}^{r-1}_{\g}\mathrm{d}h),  \quad (r \in \mathbb{Z}_+),
\end{equation}
with the initial case being $\dot{\cL}^0_{\g}\mathrm{d}h := \mathrm{d}h$. Likewise, the augmented Lie derivative can also act upon codistributions. \chang{The above geometric tools will be used for the \textit{observability} analysis by checking the observability rank condition \cite{isidori1985nonlinear}. Specifically, the observability codistribution $\Xi^\ast$ can be recursively computed according to Algorithm \ref{algorithm3}.
\begin{figure}[ht]
\removelatexerror
\begin{algorithm}[H]
\small
\vspace{-0.05cm}
\caption{Computation of the Observability Codistribution}
\label{algorithm3}
\KwData{$h_i$, $\g(\x, \uu, t)$}
\KwResult{$\Xi^\ast_i$}
\chang{
$k \longleftarrow 0$, $\text{ID} \longleftarrow 1$\; 
$\Xi^{(k)} = \text{span}\{\mathrm{d}h_1, \mathrm{d}h_2, \cdots, \mathrm{d}h_p\}$\;
\While{ID = 1}
{
$\Xi^{(k+1)} = \text{span}\{\Xi^{(k)}, \dot{\cL}_{\g}\Xi^{(k)}\}$\;
\If{$\Xi^{(k+1)} = \Xi^{(k)}$}{
$\text{ID} \longleftarrow 0$
}
$k \longleftarrow k+1$
}
\boxed{\Xi^\ast_i \longleftarrow \Xi^{(k)}_i}
}
\end{algorithm}
\end{figure}
Moreover, we have $\Xi^{(k)} \subseteq \Xi^{(k+1)}$, and the final \textit{converged} codistribution is obtained for some $k^\ast \leq n-1$ \cite{martinelli2022extension, isidori1985nonlinear}. Checking the observability rank condition is equivalent to checking the rank of the observability matrix $\OB^{(k)}$ derived from the codistribution $\Xi^{(k)}$, and $\OB^{(k)}$ is given as
\begin{equation}
    \label{eq:ob_general}
    \OB^{(k)} = 
    \begin{bmatrix}
        \textstyle\vect^{p}_{i=1}\{ \dot{\cL}^0_{\g}\mathrm{d}h_i \} \\
        \textstyle\vect^{p}_{i=1}\{ \dot{\cL}^1_{\g}\mathrm{d}h_i \} \\
        \vdots \\
        \textstyle\vect^{p}_{i=1}\{ \dot{\cL}^k_{\g}\mathrm{d}h_i \}
    \end{bmatrix},
\end{equation}
where we only refer to the coefficients of the bases $\{\mathrm{d}x_{1}, \mathrm{d}x_{2}, \cdots, \mathrm{d}x_{n}\}$ \guido{of the covector field} while the bases themselves are eliminated with slight abuse of notation. If $\text{rank}(\OB^{(k)}) = n$ for some $k$ at $(\x^\ast, t^\ast)$ with control being $\uu^\ast$, then the system is \textit{weakly locally observable} at $(\x^\ast, t^\ast)$ \cite{martinelli2005observability} \guido{with control being $\uu^\ast$}.
\begin{remark}
    In this work, we focus on \textit{active} state estimation \guido{where we also care how control inputs affect the observability of the system}. Besides, the controls $\uu$ are treated as \textit{constant parameters} in that we check observability only \textit{locally} (i.e., we assume that the controls remain constant within any small time interval). In addition, though the controls $\uu$ are treated as parameters, the space we investigate is $\uM$ in terms of active state estimation, and we aim to identify possible observable and unobservable subspaces through observability analysis.
\end{remark}
\begin{remark}
    The observability analysis only helps to determine whether the internal state of the system as in \eqref{eq:sys_math} can be \textit{well reconstructed} \guido{(i.e., the estimation error is exponentially bounded in means square and bounded with probability 1 \cite{reif1999stochastic})} by the filter that can be viewed as a \textit{stochastic realization} (e.g., the Kalman filter and its variations) \cite{van2021control}, whereas the performance of the filter (i.e., estimation accuracy, transient behavior) still depends on many other factors.
\end{remark}
}

%% file: content_file/3.3ex-Observability.tex
\subsection{Observability Analysis}
For the basic peer-to-peer \chang{localization model}, \chang{we only have a single direct distance measurement.} Noting the dynamics of the relative heading $\psi_{ij}$ is \guido{independent} of the position $\p_{ij}$ (cf. \ref{eq:dy_aug_compact}), \chang{we thus only consider the position dynamics given that $\psi_{ij}$ is also not our \guido{state variable of interest}.} After decoupling $\psi_{ij}$, the position dynamics should be treated as a TV \sven{nonlinear system} with $\psi_{ij}(t)$ being the varying parameter. \chang{\guido{Following \eqref{eq:dy_pair_compact},} the resulting peer-to-peer relative dynamics and augmented dynamics are given in \eqref{eq:sub_position_dynamics} and \eqref{eq:sub_position_augdynamics}, respectively.}
\begin{equation}
    \label{eq:sub_position_dynamics}
    \left\{
    \begin{aligned}
        \p_{ij} &= \g(\p_{ij}, \uu_{ij}, t) = \bR(\psi_{ij}(t))\vv_j - \vv_i - \dot{\psi}_i S\p_{ij} \\
        y &= h(\p_{ij}) = \|\p_{ij}\|
    \end{aligned}
    \right. ,
\end{equation}
\begin{equation}
    \label{eq:sub_position_augdynamics}
    \left\{
    \begin{aligned}
        \bar{\p}_{i} &= \bar{\g}(\bar{\p}_{i}, \bar{\uu}_{i}, t) \\
            \y^\mathrm{d}_i
                &=
            \hh^\mathrm{d}_i(\bar{\p}_i) \\
            \big(\y^\mathrm{id}_i
                &=
            \hh^\mathrm{id}_i(\bar{\p}_i)\big)
    \end{aligned}
    \right. ,
\end{equation}
where $\bar{\p}_{i} = \textstyle\vect^{N_i}_{\alpha=1}\{\p_{ij_\alpha}\}$, $\bar{\uu_i}$ is given as in \eqref{eq:aug_control},  $\bar{\g}(\bar{\p}_{i}, \bar{\uu}_{i}, t) = \textstyle\vect^{N_i}_{\alpha=1}\{\g(\p_{ij_\alpha}, \uu_{ij_\alpha}, t)\}$, and $\y^\mathrm{d}_i$ and $\y^\mathrm{id}_i$ are constructed similarly as in \eqref{eq:fcrl} (with details given in \eqref{eq:dis_direct}, \eqref{eq:dis_indirect_neighborwise} and \eqref{eq:dis_indirect_all}). We start with analyzing the peer-to-peer system given in \eqref{eq:sub_position_dynamics}, and the observability matrix $\OB^{\mathrm{d}}(i,j)$ is given as
\begin{equation}
\small{
    \label{eq:ob_matrix_pair}
    \hspace{-0.35cm}
    \OB^{\mathrm{d}}(i,j) \lsdeng
    \begin{bmatrix}[2.0]
    \begin{array}{c}
        \p^\top_{ij} \\
        \hline
        \big[\bR(\psi_{ij}(t))\vv_j - \vv_i\big]^\top \\
        \hline
        \dot{\psi}_{ij}\vv^\top_j E-\big[\dot{\psi}_j\bR(\psi_{ij}(t))\vv_j - \dot{\psi}_i \vv_i \big]^\top S
    \end{array}
    \end{bmatrix},\hspace{-0.25cm}
    }
\end{equation}
where \chang{$E = \diag\{0,0,1\}$}, \guido{$S$ is given as in \eqref{eq:dy_pair_compact}}, and all \guido{entries} are multiplied by the non-zero term $\|\p_{ij}\|$ (i.e., two agents do not collide), which does not affect the rank of $\OB^{\mathrm{d}}(i,j)$. \guido{Explicitly rewriting the above matrix lead to a more detailed expression of $\OB^{\mathrm{d}}(i,j)$ as}
\begin{equation}
\small{
    \label{eq:ob_explicit}
    \begin{bmatrix}
    \begin{array}{c|c|c}
        x_{ij} & y_{ij} & z_{ij} \\
        \hline
        \makecell*[c]{c(\psi_{ij})v_{x,j}- \\ s(\psi_{ij})v_{y,j} - v_{x,i}} 
        & \makecell*[c]{s(\psi_{ij})v_{x,j}+ \\ c(\psi_{ij})v_{y,j} - v_{y,i}} 
        &   v_{z,j} - v_{z,i} \\
        \hline
        \makecell*[c]{-\dot{\psi}_j s(\psi_{ij})v_{x,j} \\ -\dot{\psi}_j c(\psi_{ij})v_{y,j} \\ + \dot{\psi}_i v_{y,i}} 
        & \makecell*[c]{\dot{\psi}_j c(\psi_{ij})v_{x,j} \\ -\dot{\psi}_j s(\psi_{ij})v_{y,j} \\ - \dot{\psi}_i v_{x,i}} 
        &   v_{z,j} (\dot{\psi}_j - \dot{\psi}_i)
    \end{array}
    \end{bmatrix}.\notag 
    }
\end{equation}
We further denote the relative velocity $\bR(\psi_{ij}(t))\vv_j - \vv_i$ as $\vv_{ij} = [v_{x,ij}, v_{y,ij}, v_{z,ij}]^\top$. Though a complete analysis of the observability is unattainable since \guido{finding all conditions under which the determinant of the observability matrix is $0$ is difficult}, we can still find some \textit{unobservable} subspaces which have clear physical interpretation:
\begin{figure}[h]
    \centering
    \includegraphics[width=0.9\linewidth]{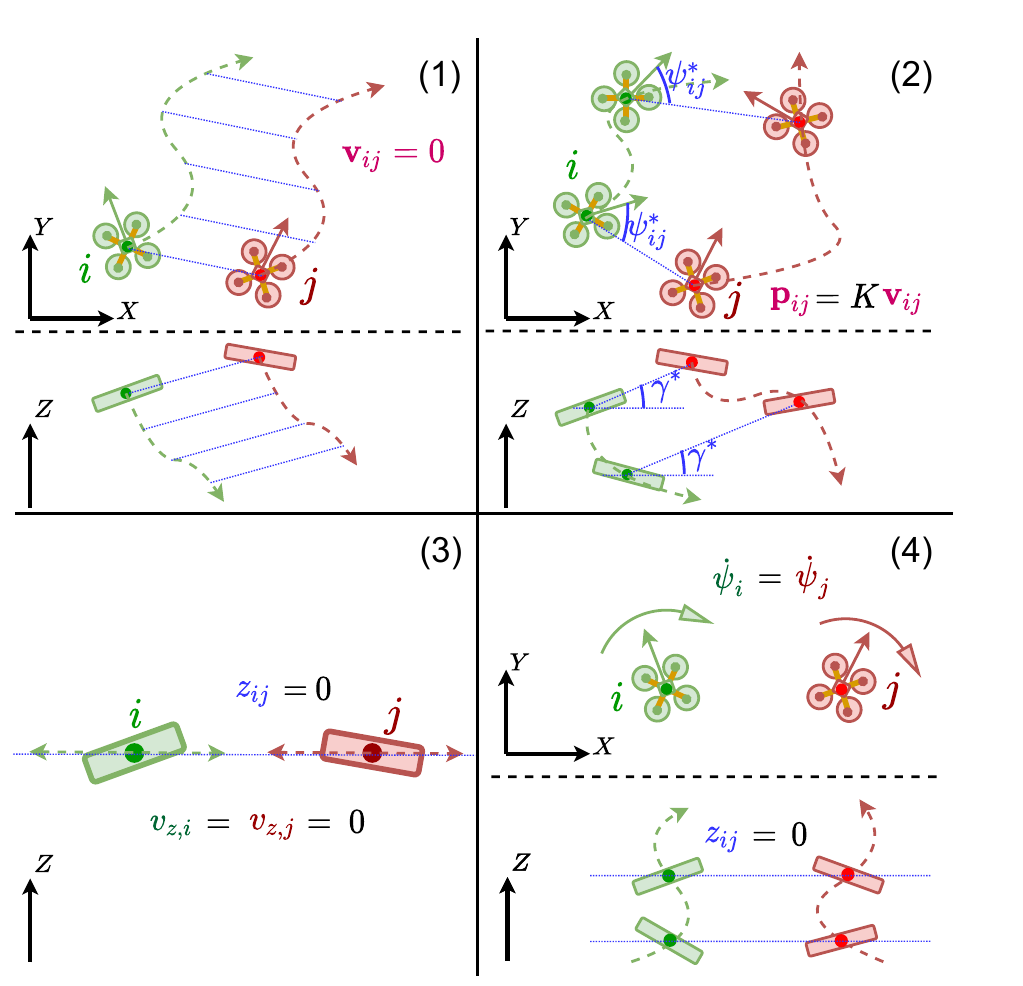}
    \caption{Unobservable motions: case (1) (top left); case (2) (top right); case (3) (bottom left); case (4) (bottom right).}
    \label{fig:observable}
\end{figure}
\begin{enumerate}
    \item Relative velocity is zero (i.e., $\vv_{ij} = \mathbf{0}$, \textbf{parallel motion}): $\OB^{\mathrm{d}}(i,j)[2] = \mathbf{0}$.
    \item Relative velocity is \textbf{aligned} with the relative position (i.e., \chang{the \textbf{relative motion is linear}}): $\OB^{\mathrm{d}}(i,j)[1] = K\OB^{\mathrm{d}}(i,j)[2] (K \neq 0)$.
    \item \chang{Two agents move in a \textbf{fixed horizontal plane} ($z_{ij} = 0$), plus that the two agents $i$ and $j$ \textbf{do not have vertical velocity} ($v_{z,i} = v_{z,j} = 0$): $\OB^{\mathrm{d}}(i,j)^\top[3] = \mathbf{0}$.}
    \item \chang{Two agents move in a \textbf{fixed horizontal plane} ($z_{ij} = 0$), plus that the two agents have the \textbf{same heading rate} ($\dot{\psi}_j - \dot{\psi}_i = 0$): $\OB^{\mathrm{d}}(i,j)^\top[3] = \mathbf{0}$.}
\end{enumerate}

The above observability analysis directly suits the nCRL model due to its dynamics being fully decoupled for each agent pair $(i,j_\alpha), j_\alpha \in \mathcal{N}_i$. 

Regarding the hCRL model, though $\uu_i$ perturbs the dynamics of each agent pair in an integrated way, both the dynamics $\g^c(\p_{ij_\alpha}, \uu_{ij_\alpha}, t)$ and the measurement model $h_{ij_\alpha} = \|\p_{ij_\alpha}\|$ are decoupled in terms of the state $\x_{ij} (\p_{ij})$. Correspondingly, the resulting observability matrix for the hCRL has a \textit{diagonal} structure as $\OB_{\text{hCRL}} = \textstyle\diag^{N_i}_{\alpha=1}\{\OB^{\text{d}}(i,j_\alpha)\}$ \chang{after some row permutations}, from which we can easily conclude that the observability for the hCRL model can be analyzed in a \textit{pair-wise} manner, \guido{making it the} same as that for the nCRL model. 

For fCRL model where indirect measurements are also present, the observable subspace can be expanded. Particularly, if the number of indirect measurements $p_{\text{id}} \geq 2N_i$ (i.e., each neighbor provides at least $2$ indirect measurements on average \guido{and these measurements do not coincide}), the total number of measurements (i.e., counting both direct and indirect ones) will be greater than the number of interested states $3N_i$, then the fCRL system is \textit{fully} observable since \chang{$\OB_{\text{fCRL}}^{(0)} = \textstyle\vect^{3N_i}_{i=1}\{\mathrm{d}h_i\}$ is already of full rank given that the agents do not collide.} \guido{The minimum number of neighboring agents that can achieve fully observable is $N_i = 5$.} In \chang{generic} cases where $p_{\text{id}} < 2N_i$, more detailed investigation is required, and we start with the simplest case where we deal with the 3 agent tuple $(i, j_1, j_2)$ without loss of generality. After some row permutations, the observability matrix $\OB(i,j_1,j_2)$ is expressed as
\begin{equation}
\small{
    \label{eq:ob_3agentfinal}
    \OB(i,j_1,j_2) = 
    \begin{bmatrix}[2]
        \frac{\OB^{\text{d}}(i,j_1)}{\|\p_{ij_1}\|} & \mathbf{0}_{3 \times 3} \\
        \mathbf{0}_{3 \times 3} & \frac{\OB^{\text{d}}(i,j_2)}{\|\p_{ij_2}\|} \\
        \frac{\OB^{\text{id}}(j_1,j_2)}{\|\p_{ij_1} - \p_{ij_2}\|} & -\frac{\OB^{\text{id}}(j_1,j_2)}{\|\p_{ij_1} - \p_{ij_2}\|}
    \end{bmatrix},
    }
\end{equation}
where the partial observability matrix $\OB^{\text{id}}(j_1,j_2)$ corresponding to the indirect measurement $h(\p_{ij_1}, \p_{ij_2}) = \|\p_{ij_1} - \p_{ij_2}\|$ is given as
\begin{equation}
\small{
    \label{eq:ob_indirect_detail}
    \hspace{-0.35cm}
    \OB^{\text{id}}(j_1,j_2) \lsdeng
    \begin{bmatrix}[2.5]
    \begin{array}{c}
        \p^\top_{ij_1} - \p^\top_{ij_2} \\
        \hline
        [R(\psi_{ij_1})\vv_{j_1} - R(\psi_{ij_1})\vv_{j_2}]^\top \\
        \hline
        \makecell[c]{
        (\dot{\psi}_{ij_1}\vv_{j_1} - \dot{\psi}_{ij_2}\vv_{j_2})^\top E
        \\
        -[\dot{\psi}_{ij_1}R(\psi_{ij_1})\vv_{j_1} - \dot{\psi}_{ij_2}R(\psi_{ij_2})\vv_{j_2}]^\top S
        }
    \end{array}
    \end{bmatrix},\hspace{-0.25cm}
    }
\end{equation}
\chang{We then discuss how the observable subspace can be expanded by imposing extra stringent conditions on the previously identified unobservable subspace. In the following analysis, we only discuss the cases where \guido{the previously identified $4$ unobservable conditions hold independently}, whereas the cases in which multiple conditions are satisfied simultaneously will not be further investigated.} \guido{In the following, we will frequently use the term \textit{scaled velocity} which means the velocity scaled by the heading rate (e.g., $\dot{\psi}_j\vv_j$ and $\dot{\psi}_j\bR(\psi_{ij})\vv_j$), and the scaled velocity terms are present in the observability subblocks (cf. \eqref{eq:ob_matrix_pair} and \eqref{eq:ob_indirect_detail}).}
\begin{enumerate}
    \item Parallel motion
    \begin{itemize}
        \item \chang{If either $(i,j_1)$ or $(i,j_2)$ is in parallel motion, then the system is always observable in that $\OB^{\text{id}}(j_1,j_2)[1] \neq 0$ always holds true.}
        \item \chang{If both $(i,j_1)$ and $(i,j_2)$ are in parallel motion (i.e., all 3 agents are in parallel motion), then the system is still observable if the relative scaled velocity of pair $(j_1, j_2)$ is non-zero (i.e., $\OB^{\text{id}}(j_1,j_2)[3] \neq 0$)}
    \end{itemize}
    \item Relative velocity aligned with the relative position
    \begin{itemize}
        \item \chang{If either $(i,j_1)$ or $(i,j_2)$ falls in this aligning condition, then the system is always observable in that $\OB^{\text{id}}(j_1,j_2)[1] \neq 0$ always holds true.}
        \item \chang{If both pair $(i,j_1)$ and $(i,j_2)$ satisfy this aligning condition, then the system is still observable if pair $(j_1, j_2)$ does not meet this condition. Further, even if $(j_1, j_2)$ also satisfies this aligning condition, then the system is still observable if the relative scaled velocity of pair $(j_1, j_2)$ is non-zero (i.e., $\OB^{\text{id}}(j_1,j_2)[3] \neq 0$).} 
    \end{itemize}
    \item Fixed horizontal plane plus zero vertical velocity \\
    \chang{If at least one of the two pairs $(i, j_1)$ and $(i, j_2)$ satisfies the condition that $i$ and $j_1$ ($j_2$) moves in the same horizontal plane and $j_1$ ($j_2$) does not have vertical velocity, then the system is still observable if any one of the below conditions holds: i) $j_1$ and $j_2$ have altitude difference ii) $j_1$ and $j_2$ have different vertical velocity iii) the relative scaled vertical velocity of pair $(j_1, j_2)$ is non-zero \guido{(i.e., $\OB^{\text{id}}(j_1,j_2)[3] \neq 0$)}.}
\end{enumerate}

\chang{For the general fCRL model of agent $i$ and its neighbor set $\mathcal{N}_i$, its observability matrix $\OB(i,\mathcal{N}_i)$ is given as
\begin{equation}
\def\arraystretch{2}
    \label{eq:ob_fcrl_general}
    \begin{bNiceMatrix}[first-row, last-col, respect-arraystretch]
        \mathrm{d}\p_{ij_1} & \mathrm{d}\p_{ij_2} & \cdots & \mathrm{d}\p_{ij_{N_i}} & \\
        \frac{\OB^{\text{d}}(i,j_1)}{\|\p_{ij_1}\|} & \mathbf{0}_{3 \times 3} & \cdots & \mathbf{0}_{3 \times 3} & d_{ij_1}\\
        \mathbf{0}_{3 \times 3} & \frac{\OB^{\text{d}}(i,j_2)}{\|\p_{ij_2}\|} & \cdots & \mathbf{0}_{3 \times 3} & d_{ij_2}\\
        \vdots & & \ddots & \vdots & \vdots \\
        \mathbf{0}_{3 \times 3} & \cdots & & \frac{\OB^{\text{d}}(i,j_{N_i})}{\|\p_{ij_{N_i}}\|} & d_{ij_{N_i}} \\
        \hline
        \frac{\OB^{\text{id}}(j_1,j_2)}{\|\p_{ij_1} - \p_{ij_2}\|} & -\frac{\OB^{\text{id}}(j_1,j_2)}{\|\p_{ij_1} - \p_{ij_2}\|} & \cdots & \mathbf{0}_{3 \times 3} & d_{j_1j_2}\\
        \frac{\OB^{\text{id}}(j_1,j_{N_i})}{\|\p_{ij_1} - \p_{ij_{N_i}}\|} & \mathbf{0}_{3 \times 3} & \cdots & -\frac{\OB^{\text{id}}(j_1,j_{N_i})}{\|\p_{ij_1} - \p_{ij_{N_i}}\|} & d_{j_1j_{N_i}} \\
        \vdots & \vdots & \vdots & \vdots & \vdots
    \end{bNiceMatrix}, \notag
\end{equation}
}
where \guido{the upper part with only diagonal subblocks} is fixed and the lower part depends on the type of indirect measurements. Specifically, the direct measurement $d_{ij_\alpha}$ and the indirect measurement $d_{j_\alpha j_\beta}$ contribute, respectively, to the following two blocks:
\begin{equation}
\small{
    \label{eq:ob_indi_detial}
    \left \{
    \begin{aligned}
        d_{ij_\alpha}: \;& \mathbf{b}^{N_i}_{\alpha} \otimes \frac{\OB^{\text{id}}(j_i,j_{\alpha})}{\|\p_{ij_\alpha}\|} \\
        d_{j_\alpha j_\beta}: \;& (\mathbf{b}^{N_i}_{\alpha} - \mathbf{b}^{N_i}_{\beta}) \otimes \frac{\OB^{\text{id}}(j_\alpha,j_{\beta})}{\|\p_{ij_\alpha} - \p_{ij_{\beta}}\|}
    \end{aligned}
    \right. .}
\end{equation}
Based on the special structure of the observability matrix, the general rule of thumb is that the more indirect measurements that agent $j_\alpha$ is coupled, the more the unobservable subspace corresponding to the state $\p_{ij_\alpha}$ diminishes.

%% file: content_file/4-main_approaches.tex
\section{\chang{Filtering Based State Estimation}}
\label{sec:4-mainapp}
This section provides the nonlinear filtering solution to the state estimation problem in CRL. We will give an introduction to the basic extended Kalman filter (EKF), with which we develop the MLVC-EKF and apply it to the CRL models. The convergence analysis of the fixed-point iteration involved in the posterior estimation will also be presented. Finally, we analyze the computational complexity of all 3 CRL schemes applied with both MLVC-EKF and the EKF.

\input{content_file/4.1-Elementary-EKF.tex}
\input{content_file/4.2-MLVC-EKF.tex}
\input{content_file/4.3-Fixed-point.tex}
\input{content_file/4.4-Complexity.tex}

%% file: content_file/4.1-Elementary-EKF.tex
\subsection{\sven{Extended Kalman Filter}}
We first summarize the basic extended Kalman filter algorithm. To better serve our built CRL models, we consider \guido{a continuous-time stochastic nonlinear} system of the form
\begin{equation}
\label{eq:sys_filter}
\left \{
\begin{aligned}
    \dot{\x} &= \g(\x, \uu, \ww) \\
    \y &= \hh(\x) + \NU
\end{aligned}
\right. ,
\end{equation}
where $\x \in \mathcal{M} \subseteq \mathbb{R}^n$ is the state, $\y \in \mathcal{Y} \subseteq \mathbb{R}^p$ is the \guido{measurement vector}, $\uu \in \mathcal{U} \subseteq \mathbb{R}^m$ is the control input, and $\ww \in \mathcal{W} \subseteq \mathbb{R}^m$ and $\NU \in \mathcal{V} \subseteq \mathbb{R}^p$ are process noise and measurement noise, respectively. The system evolves on an open time interval $\mathcal{I}$. With sampling time $T_s$, the continuous-time system can be discretized using \guido{the Euler method and evaluated} at $t_0 + kT_s\;(k \in \mathbb{Z}^0_{k_\text{m}})$ such that $\{t_0 + kT_s\}^{k_\text{m}}_{k=0} \subset \mathcal{I}$. We simply use $k$ as the index, and the corresponding discrete-time system is given as
\begin{equation}
\label{eq:sys_filter_discrete}
\left \{
\begin{aligned}
    \x_{k} &= \x_{k-1} + T_s\g(\x_{k-1}, \uu_{k-1}, \ww_{k-1}) \\
    \y_{k} &= \hh(\x_{k}) + \NU_{k}
\end{aligned}
\right. .
\end{equation}

We then provide the EKF \textit{recursion} for the discretized system as in \eqref{eq:sys_filter_discrete}. Given the corrected estimated state $\hbx^+_{k-1}$ as well as its covariance $\bP^+_{k-1}$ at time instant $k-1 (k \geq 1)$, we then proceed to compute the posterior estimation $\hbx^+_{k}$ and covariance $\bP^+_{k}$ \sven{in two steps}.

\noindent \textbf{1) Prediction}: Given the control input $\uu_{k-1}$, compute the prior estimate $\hbx^-_{k}$ and its covariance $\bP^-_{k}$ as
\begin{subequations}
\label{eq:ekf_prediction}
    \begin{align}
        \label{eq:ekf_prediction_state}
        \hbx_{k}^- &= \hbx_{k-1}^+ + T_s\g(\hbx^+_{k-1}, \uu_{k-1}, \mathbf{0})\\
        \label{eq:ekf_prediction_var}
        \bP^-_{k} &= \A_{k-1}\bP^+_{k-1}\A^\top_{k-1} + \B_{k-1}\mathbf{Q}_{k-1}\B^\top_{k-1},
    \end{align}
\end{subequations}
where $Q_{k-1} = \hat{\mathbb{E}}[\ww_{k-1}\ww^\top_{k-1}]$ is the estimated process noise covariance matrix, and $\A_{k-1}$ and $\B_{k-1}$ are given as
\begin{subequations}
\small{
\label{eq:dynamics_linearization}
    \begin{align}
        \label{eq:dynamics_linearization_A}
        \A_{k-1} &= \chang{\mathds{I}_{n} + T_s \bigg(\frac{\partial \g(\x, \uu, \ww)}{\partial \x}\bigg|_{(\hbx^+_{k-1}, \uu_{k-1}, \mathbf{0})} \bigg)},\\
        \label{eq:dynamics_linearization_B}
        \B_{k-1} &= \chang{T_s \bigg(\frac{\partial \g(\x, \uu, \ww)}{\partial \uu}\bigg|_{(\hbx^+_{k-1}, \uu_{k-1}, \mathbf{0})} \bigg)}.
    \end{align}
}
\end{subequations}

\noindent \textbf{2) Correction by measurements}: Correct the prior estimation based on the measurement $\y_{k}$, which outputs the \textit{posterior} estimation $\hbx^+_{k}$ and its covariance $\bP^+_{k}$. First, approximate $\y_{k}$ using first-order linearization as
\begin{equation}
    \label{eq:meas_approximation}
    \y_{k} \approx \hh(\hbx^-_{k}) + \bH_{k}(\x_{k} - \hbx^-_{k}) + \NU_{k},
\end{equation}
where $\bH_{k}$ is given as
\begin{equation}
\small{
    \label{eq:meas_linearization}
    \bH_{k} = \frac{\partial \hh(\x)}{\partial \x}\bigg|_{\hbx^-_{k}}.
    }
\end{equation}
\chang{
Then the Kalman update follows as
\begin{subequations}
    \label{eq:basic-kalman_update_general}
    \begin{align}
    \label{eq:basic-kalman_update_general-state}
        \hbx^+_{k} &= \hbx^-_{k} + \K_k(\y_{k} - \hh(\hbx^-_{k})), \\
    \label{eq:basic-kalman_update_general-var}
        \bP^{+}_{k} &= (\mathds{I}_n \lsjian \K_k\mathbf{H}_{k})\bP^-_{k}(\mathds{I}_n \lsjian \K_k\mathbf{H}_{k})^{\top}+\K_k\bR_k\K_k{^\top},
    \end{align}
\end{subequations}
where $\bR_k = \hat{\mathbb{E}}[\NU_{k}\NU^\top_{k}]$ is the estimated measurement noise covariance matrix, and the Kalman gain $\K_k$ is given as
\begin{equation}
    \label{eq:basic-kalman_gain}
    \K_k = \bP_{k}^-\bH_k^\top\big(\bH_{k}\bP_{k}^-\bH^\top_{k} + \bR_{k}\big)^{-1}.
\end{equation}
}

%% file: content_file/4.2-MLVC-EKF.tex
\subsection{\sven{Kernel-induced Extended Kalman Filter}}
\begin{figure}
    \centering
    \includegraphics[width=\linewidth]{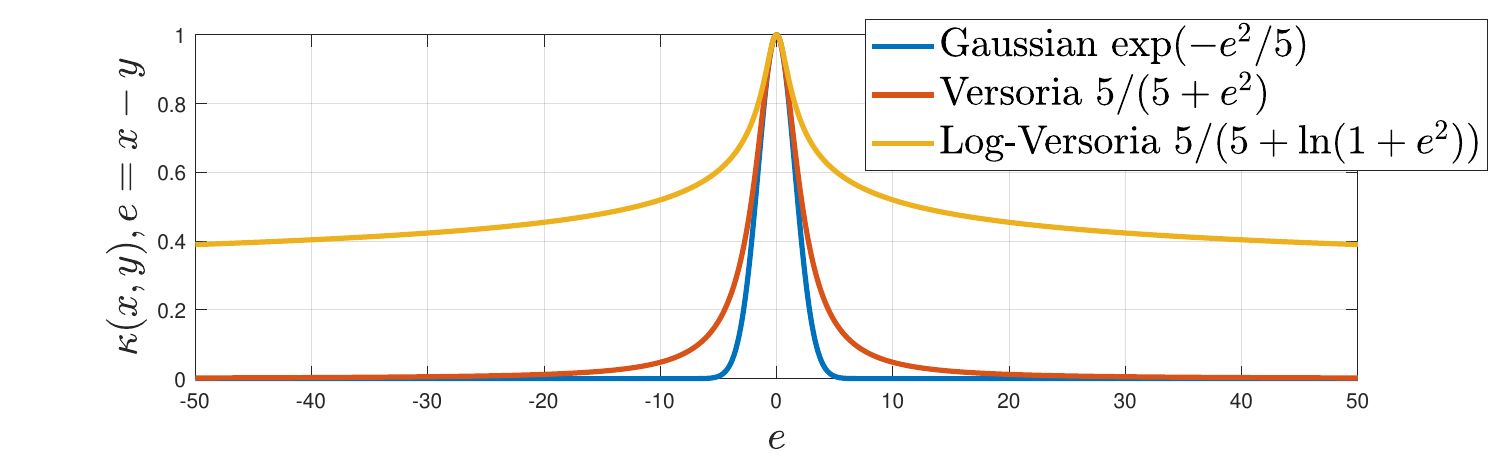}
    \caption{Kernel Function Comparison}
    \label{fig:kernel}
\end{figure}

\guido{To cope with heavy-tailed non-Gaussian measurement noise, we use correntropy as a new metric to define the estimation error.} In information theoretical learning, correntropy measures the similarity between two random variables $X, Y \in \mathbb{R}$ as
\begin{equation}
    \label{eq:corr_definition}
    C(X,Y) = \mathbb{E}[\kappa(X,Y)] = \int \kappa(x,y)\mathrm{d}F_{XY}(x,y),
\end{equation}
where $F_{XY}(x,y)$ is the joint distribution function and $\kappa(\cdot,\cdot)$ is a shift-invariant Mercer kernel \guido{\cite{principe2010information}}. In this work, inspired by the Versoria function\cite{huang2017maximum}, we design a novel Logarithmic-Versoria (LV) kernel as
\begin{equation}
    \label{eq:ker_definition}
    \kappa(x,y) = L_{\tau}(e) = \frac{\tau}{\tau + \ln(1 + e^2)},
\end{equation}
where $e=x-y$ is the error, and $\tau > 0$ controls the \textit{bandwidth} of the kernel. The Versoria function can exploit the non-Gaussian characteristics and provide better higher-order error information than the Gaussian kernel \cite{huang2017maximum}. The newly designed LV kernel not only inherits the power of Versoria function but also exhibits smoother behavior for small errors. \chang{A brief visualization of different kernel functions is given in Figure \ref{fig:kernel}, \guido{where we compare the Gaussian kernel function, the Versoria kernel function, and the Logarithmic-Versoria function all with the same bandwidth $\tau = 5$ (cf. \eqref{eq:gaussian_kernel}, \eqref{eq:versoria_kernel}, and \eqref{eq:ker_definition}).}}

For the filtering problem of general stochastic systems where the analytical distribution of the state(output) is not \guido{known}, it is practical to approximate the correntropy using the sampled-mean strategy as
\chang{\begin{equation}
    \label{eq:corr_sample}
    \hat{C}(X,Y) = \frac{1}{D}\sum^{D}_{i=1}L_{\tau}(e(i)),
\end{equation}}
where $e(i)=x(i)-y(i)$ with $\{x(i),y(i)\}$ being the candidate data points. 

\chang{The MLVC-EKF \textit{recursion} also has the two-step structure as the standard EKF, and it only differs from the EKF in the state correction step whereas the prediction step exactly follows \eqref{eq:ekf_prediction_state}. The details of the correction step in the MLVC-EKF is presented as follows.}

\noindent \textbf{\chang{2) Kernel-induced correction}}: \chang{Based on the linearized measurement equation in \eqref{eq:meas_linearization}, we then leverage both the state error $\mathbf{e}_{\x,k}:= \hbx^-_{k} -\x_{k} $ and the measurement noise $\NU_{k}$ to obtain}
\begin{equation}
    \label{eq:error_augmentation}
    \hspace{-0.2cm}
    \underbrace{
    \begin{bmatrix}
    \hbx^-_{k} \\
    \y_{k} - \hh(\hbx^-_{k}) + \bH_{k}\hbx^-_{k}
    \end{bmatrix}}_{:=\mathbf{z}_{k}}
    \hspace{-0.1cm}
    =
    \hspace{-0.1cm}
    \underbrace{
    \begin{bmatrix}
    \mathds{I}_{n} \\
    \bH_{k}
    \end{bmatrix}}_{:=\mathbf{F}_{k}}
    \x_{k}
    +
    \underbrace{
    \begin{bmatrix}
    \mathbf{e}_{\x,k}  \\
    \NU_{k}
    \end{bmatrix}}_{:=\bm{\delta}_{k}}
    \hspace{-0.13cm}.
\end{equation}
To normalize the error $\bm{\delta}_{k}$, we perform Cholesky decomposition to obtain $\M_{\x,k} = \CH(\bP^-_{k})$ and $\M_{\y,k} = \CH(\bR_k)$. We further define $\M_k = \diag\{\M_{\x,k}, \M_{\y,k}\}$, then \eqref{eq:error_augmentation} can be modified as 
\begin{equation}
    \label{eq:error_aug_modified}
    \underbrace{(\M_k)^{-1}\mathbf{z}_{k}}_{:=\mathbf{z}^\ast_{k}}
    = \underbrace{(\M_k)^{-1}\mathbf{F}_{k}}_{:=\mathbf{F}^\ast_{k}} \x_{k}
    + (\M_k)^{-1}\bm{\delta}_{k}.
\end{equation}
For the regression model given in \eqref{eq:error_aug_modified}, we incorporate the Versoria loss function (cf. \eqref{eq:corr_sample}) as
\begin{equation}
    \label{eq:loss_function}
    \hspace{-0.2cm}
    J(\x_{k}) = \frac{1}{D}\sum^{D}_{i=1}L_{\tau}\bigg(
    \underbrace{\mathbf{z}^\ast_{k}[i] -\mathbf{F}^\ast_{k}[i,:]\x_{k}}_{:=\mathbf{e}_{k}[i]}
    \bigg),
\end{equation}
where \guido{$D := n + p$} is the dimension of the regression problem, with which the posterior estimation $\hbx^+_{k}$ can be computed by solving the following optimization problem:
\begin{equation}
    \label{eq:opt_posteriori}
    \hbx^+_{k} = \arg\max_{\x_{k}}J(\x_{k}).
\end{equation}

The optimality of \eqref{eq:opt_posteriori} is guaranteed by $\partial J(\x_{k})/ \partial \x_{k} = 0$, which suggests that $\hbx^+_{k}$ satisfies the following equation:
\begin{equation}
    \label{eq:equation_optimality}
    \hbx^+_{k} = \big({\bF^\ast_k}^\top \LL_k {\bF^\ast_k}\big)^{-1}{\bF^\ast_k}^\top \LL_k \mathbf{z}^\ast_{k},
\end{equation}
where $\LL_k = \textstyle\diag\{\LL_{\x,k}, \LL_{\y,k}\}$ with $\LL_{\x,k}$ and $\LL_{\y,k}$ being constructed as follows:
\begin{equation}
    \label{eq:kernel_matrix}
    \left \{
    \begin{aligned}
        \LL_{\x,k} &= \textstyle\diag^n_{i=1}\big\{L_{\tau}^2(\hat{\mathbf{e}}_k[i]) (1+\hat{\mathbf{e}}^2_k[i])^{-1}\big\} \\
        \LL_{\y,k} &= \textstyle\diag^{D}_{i=n+1}\big\{L_{\tau}^2(\hat{\mathbf{e}}_k[i])(1+\hat{\mathbf{e}}^2_k[i])^{-1}\big\} \\
    \end{aligned}
    \right. .
\end{equation}
In \eqref{eq:kernel_matrix}, we have $\hat{\mathbf{e}}_k[i] = \mathbf{z}^\ast_{k}[i] -\mathbf{F}^\ast_{k}[i,:]\hbx^+_{k}$. Moreover, \eqref{eq:equation_optimality} can be rewritten as a Kalman-type state update \cite{chen2017maximum}
\begin{equation}
    \label{eq:kalman_update_general}
    \hbx^+_{k} = \hbx^-_{k} + \K_k(\y_{k} - \hh(\hbx^-_{k})),
\end{equation}
where the Kalman gain $\K_k$ is given as
\begin{equation}
    \label{eq:kalman_gain}
    \K_k = \bP_{\LL,k}^-\bH_k^\top\big(\bH_{k}\bP_{\LL,k}^-\bH^\top_{k} + \bR_{\LL,k}\big)^{-1},
\end{equation}
where the \textit{kernel-weighted} covariance matrices $\bP_{\LL,k}^-$ and $\bR_{\LL,k}$ are given as
\begin{equation}
    \label{eq:kernel_wighted_variance}
    \left \{
    \begin{aligned}
        \bP_{\LL,k}^- = \M_{\x,k}\LL_{\x,k}^{-1}\M^\top_{\x,k} \\
        \bR_{\LL,k} = \M_{\y,k}\LL_{\y,k}^{-1}\M^\top_{\y,k}
    \end{aligned}
    \right. .
\end{equation}
Since the right hand side of \eqref{eq:kalman_update_general} implicitly depends on $\hbx^+_{k}$, computing $\hbx^+_{k}$ requires the \textit{fixed-point} algorithm. The converged output of the fixed-point iteration will serve as the posterior estimation, and the following final step is the Kalman-type covariance update \chang{as in \eqref{eq:basic-kalman_update_general-var}}. The overall MLVC-EKF is summarized in Algorithm \ref{algorithm1}.
\begin{figure}[ht]
\removelatexerror
\begin{algorithm}[H]
\small
\vspace{-0.05cm}
\caption{MLVC-EKF State Estiamtion}
\label{algorithm1}
\KwData{$\hbx^+_{k-1}$, $\bP^+_{k-1}$,
$\uu_{k-1}$,
$\y_{k}$,
$\mathbf{Q}_k$,
$\bR_k$,
$T_s$}
\KwResult{$\hbx^+_{k}$, $\bP^+_{k}$}
Compute $\A_{k-1}$, $\B_{k-1}$ (cf. \eqref{eq:dynamics_linearization})\;
Compute $\hbx^-_{k}$, $\bP^-_{k}$ (cf. \eqref{eq:ekf_prediction}), $\mathbf{H}_{k}$ (cf. \eqref{eq:meas_linearization})\;
$\M_{\x,k} = \CH(\bP^-_{k})$, $\M_{\y,k} = \CH(\bR_k)$\;
$\text{Tag}_{\text{FP}} \longleftarrow 1$, $t \longleftarrow 0$, $\x^{(0)}_k \longleftarrow \hbx^-_{k}$, $\iota = n + 1$\;
\While{$\text{Tag}_{\text{FP}}$}
{
Build $\mathbf{z}^\ast_{k}$, $\mathbf{F}^\ast_k$ (cf. \eqref{eq:error_augmentation}, \eqref{eq:error_aug_modified})\;
$\mathbf{e}_k^{(t)} = \mathbf{z}^\ast_{k} - \mathbf{F}^\ast_k \x^{(t)}_k$\;
\chang{
$\LL^{(t)}_{\x,k} \lsdeng \textstyle\diag^n_{i=1}\{L_{\tau}^2(\hat{\mathbf{e}}^{(t)}_k[i])[1 + (\hat{\mathbf{e}}^{(t)}_k[i])^2]^{-1}\}$\;
$\bP_{\LL,k}^{(t)} \lsdeng \M_{\x,k}\LL^{(t)}_{\x,k}{^{-1}}\M^\top_{\x,k}$\;
$\LL^{(t)}_{\y,k} \lsdeng \textstyle\diag^{D}_{i=\iota}\{L_{\tau}^2(\hat{\mathbf{e}}^{(t)}_k[i])[1 + (\hat{\mathbf{e}}^{(t)}_k[i])^2]^{-1}\}$\;
$\mathbf{R}_{\LL,k}^{(t)} \lsdeng \M_{\y,k}\LL^{(t)}_{\y,k}{^{-1}}\M^\top_{\y,k}$\;
}
$\K^{(t)}_k = \bP_{\LL,k}^{(t)}\bH^\top_{k}\big(\bH_{k}\bP_{\LL,k}^{(t)}\bH^\top_{k} + \mathbf{R}_{\LL,k}^{(t)}\big)^{-1}$\;
$\x^{(t+1)}_k = \hbx^-_{k} + \K^{(t)}_k(\y_{k} - \hh(\hbx^-_{k})$\;
$r = \|(\x^{(t+1)}_k-\x^{(t)}_k)/\x^{(t)}_k\|$\;
\If{$r \leq \epsilon$}
{$\text{Tag}_{\text{FP}} \longleftarrow 0$, \boxed{\hbx^+_{k} \longleftarrow\x^{(t+1)}_k}
$\K^\ast_k\longleftarrow\K^{(t)}_k$}
}
\boxed{\bP^{+}_{k} = (\mathds{I}_n - \K^\ast_k\mathbf{H}_{k})\bP^-_{k}(\mathds{I}_n - \K^\ast_k\mathbf{H}_{k})^{\top}+\K^\ast_k\bR_k\K^\ast_k{^\top} }
\end{algorithm}
\end{figure}
\begin{remark}
The covariance update at the end of the MLVC-EKF algorithm adopts the \textit{Joseph form} to guarantee symmetry as well as positive definiteness of the covariance matrix in numerical computation, which is necessary for performing the Cholesky decomposition.
\end{remark}
\begin{remark}
The fixed-point iteration works with temporary state variable $\x^{(t)}_k$, and it terminates when the norm of the \textit{incremental percentile} ($(\x^{(t+1)}_k-\x^{(t)}_k)/\x^{(t)}_k$) is smaller than a preset threshold $\epsilon$. A smaller value of $\epsilon$ leads to a more accurate estimation but may increase the required number of fixed-point iterations. In practice, we normally choose $\epsilon$ within the range $[10^{-6}, 10^{-3}]$.
\end{remark}

Having developed the MLVC-EKF state estimation algorithm, we now investigate the application details of CRL. Due to the limited communication bandwidth, each agent can only \guido{bind} with its neighbors \textit{sequentially}. As a consequence, at each time instant, the received inputs for propagating the dynamic model and the available measurements are limited. Thus, We define the \textit{information package} $\IP(j_{\alpha},k)$ as below:
\begin{definition}[Information Package] For agent $i$ performing CRL task with its neighboring set being $\mathcal{N}_i$, the received messages from agent $j_{\alpha}$ at \guido{iteration $k$} is denoted as $\IP(j_{\alpha},k)$ which contains the following details:
\begin{itemize}
    \item $\IP(j_{\alpha},k).\uu$: The control inputs of agent $j_{\alpha}$.
    \item $\IP(j_{\alpha},k).\y$: For fCRL mode, the latest stored distances $d_{lj_{\alpha}}$ of its neighbors $l \in \mathcal{N}_{j_{\alpha}}$ and the newly measured distance $d_{ij_{\alpha}}$. For hCRL/nCRL mode, only the newly measured distance $d_{ij_{\alpha}}$.
\end{itemize}
\end{definition}
Noting the input-affine structure of the CRL model, there exists a simple CRL mechanism that only performs state estimation upon receiving the information package from all its neighbors. Compared to promptly performing estimation whenever an information package arrives, this unified scheme is more computationally efficient in that the fixed-point iteration only needs to be executed once \chang{if MLVC-EKF is applied}. With a properly designed communication protocol that ensures \textit{persistent and fair} \guido{binding} for each of the neighbors, this unified scheme is a feasible solution. 

\sven{Another issue is the package dropout which frequently occurs in practice. If the measurement $\IP(j_{\alpha},k).\y$ (partly) gets lost, then fewer measurements will be used to compute the posteriori. In extreme cases where no measurement is received from any neighbors, then the second correction step will simply be ignored whereas only prediction is carried out. Package dropout in terms of $\IP(j_{\alpha},k).\y$ is more tricky, and we need a backup algorithm that computes \textit{predicted} inputs using previously stored input data (i.e., extrapolation).}
\begin{figure}[ht]
\removelatexerror
\begin{algorithm}[H]
\small
\caption{CRL Mechanism}
\label{algorithm2}
\KwIn{fCRL(cf. \eqref{eq:fcrl})/hCRL(cf. \eqref{eq:hcrl})/nCRL(cf. \eqref{eq:hcrl}), $\mathcal{N}_i$, $\hbx_0^+$, $\bP_0^+$, $T_s$, $\{\uu_{i,k}\}^{k_m-1}_{k=0}$, $\{\IP(j_{\alpha},k)\}^{k_m}_{k=1}, \forall j_{\alpha} \in \mathcal{N}_i$ }
\KwOut{$\{\hbx^+_{k}\}^{k_m}_{k=1}$, $\{\bP^+_{k}\}^{k_m}_{k=1}$}
$k \longleftarrow 1$, $\text{mode} \longleftarrow 0$\;
\If{fCRL/hCRL}{$\text{mode} = 1$}
\While{$k \leq k_m$}
{
Build $\bar{\y}^{\mathrm{d}}_{i,k}$ by augmenting $d_{ij_\alpha}$ from $\IP(j_{\alpha},k).\y$, $\forall j_{\alpha} \in \mathcal{N}_i $\;
Build $\bar{\y}^{\mathrm{id}}_{i,k}(j_{\alpha})$ by augmenting $d_{lj_\alpha}$ from $\IP(j_{\alpha},k).\y$, $\forall l \in \mathcal{N}_i \cap \mathcal{N}_{j_{\alpha}}$\;
$\bar{\y}^{\mathrm{id}}_{i,k} \longleftarrow \textstyle\vect^{N_i}_{\alpha=1}\{\bar{\y}^{\mathrm{id}}_{i,k}(j_{\alpha})\}$
\Comment*[r]{For hCRL/nCRL, the generated $\bar{\y}^{\mathrm{id}}_{i,k} = \emptyset$}
$\uu_{j_{\alpha},k-1} \longleftarrow \IP(j_{\alpha},k).\uu, \forall j_\alpha \in \mathcal{N}_i$\;
\eIf{\guido{$\text{mode} = 1$}}
{
$\y_{k} = \vect\{\bar{\y}^{\mathrm{d}}_{i,k}, \bar{\y}^{\mathrm{id}}_{i,k}\}$\;
Compute $\bR_k$ based on $\y_{k}$\;
$\bar{\uu}_{i,k-1} \lsdeng \vect\{\uu_{i,k-1} , \textstyle\vect^{N_i}_{\alpha=1} \{\uu_{j_{\alpha},k-1})^\top \}\}$\;
$\Q_k = \mathds{I}_{N_i+1} \otimes \Q_{\uu}$\;
Compute \boxed{\hbx^+_{k}, \bP^+_{k}} by MLVC-EKF (cf. Alg \ref{algorithm1})
\Comment*[r]{Use augmented dynamics (cf.\eqref{eq:dy_aug_compact})}
}
{
Decomposition to obtain $\hbx^+_{\alpha,k-1}, \bP^+_{\alpha, k-1}$ \chang{(the inverse operation of step 25)}\;
\For {$\alpha = 1,2,\cdots,N_i$}
{
$\y_{k} \longleftarrow d_{ij_\alpha}$\;
Compute $\bR_k$ based on $\y_{k}$\;
$\uu_{ij_{\alpha},k-1} = \vect\{\uu_{i,k-1}, \uu_{j_{\alpha},k-1}\}$\;
$\Q_k = \diag\{\Q_{\uu}, \Q_{\uu}\}$\;
Compute \boxed{\hbx^+_{\alpha, k}, \bP^+_{\alpha, k}} by MLVC-EKF (cf. Alg \ref{algorithm1})
\Comment*[r]{Use peer-to-peer dynamics (cf.\eqref{eq:dy_pair_compact})}
}
\boxed{\hbx^+_{k} \lsdeng \textstyle\vect^{N_i}_{\alpha=1}\{\hbx^+_{\alpha, k}\}, \bP^+_{k} \lsdeng \textstyle\diag^{N_i}_{\alpha=1}\{\bP^+_{\alpha, k}\} }\;
}
$k \longleftarrow k + 1$
}
\end{algorithm}
\end{figure}
\begin{remark}
    Algorithm 2 is a synthesized CRL mechanism tailored for all 3 CRL models (i.e., fCRL, hCRL, and nCRL). When proceeding with the algorithm, one should simply choose one of the 3 models.
\end{remark}
\begin{remark}
    For the unified CRL scheme, the sampling time $T_s$ is the average \guido{binding} interval for a pair of agents. With a properly designed communication protocol, $T_s$ is identical for all pairs \chang{$(i, j_\alpha)$}.
\end{remark}
\begin{remark}
    For the indirect distance $d_{lj_\alpha}$ (cf. step $7$ of Algorithm \ref{algorithm2}), it will be transmitted twice by agent $j_\alpha$ and $l$ sequentially, but only the latest transmitted distance will be used as an effective measurement. 
\end{remark}
\begin{remark}
    The details of the measurement noise covariance matrix $\mathbf{R}_k$ (cf. step $12$) and the actuator noise covariance matrix $\mathbf{Q}_{\uu}$ (cf. step $14$, $22$) will be given in Section \ref{sec:5-simulation}.
\end{remark}

%% file: content_file/4.3-Fixed-point.tex
\subsection{Convergence of The Fixed-point Iteration}
The core part of the MLVC-EKF is the kernel-based measurement update which provides the posterior estimation using the fixed-point iteration. The convergence of the fixed-point iteration is thus important for the successful implementation of the filter. In the following results, we omit the time index $k$ and the superscript $+$ for simplicity, and we rewrite \eqref{eq:equation_optimality} in a compact form as
\begin{equation}
    \label{eq:fixed_point_compact}
    \hbx = \mathbf{f}(\hbx)
\end{equation}
\guido{Following \cite{chen2017maximum}}, the main results regarding the convergence of \eqref{eq:fixed_point_compact} are summarized in Theorem \ref{theorem1}.
\begin{theorem}
\label{theorem1}
Given $\rho > \xi$ (cf. \eqref{eq:xi_definition}) and $\tau \geq \max\{\tau^\ast, \tau^\dagger\}$, where $\tau^\ast$ satisfies $\Upsilon(\tau^\ast) = \rho$ (cf. \eqref{eq:tau_limits1}) and $\tau^\dagger$ satisfies $\Phi(\tau^\dagger; \rho) = \zeta\; (0 < \zeta < 1)$ (cf. \eqref{eq:tau_limits2}). Then for \sven{$\hbx \in \{\mathbf{m} \in \mathbb{R}^n | \|\mathbf{m}\|_1 \leq \rho\}$}, the following two conditions hold:
\begin{subequations}
    \label{eq:fixed_point_condition}
    \begin{align}
    \label{eq:fp_condi_f}
    &\|\mathbf{f}(\hbx)\|_1 \leq \rho, \\
    \label{eq:fp_condi_jacobian}
    &\|\nabla \mathbf{f}(\hbx)\|_1 \leq \zeta.
    \end{align}
\end{subequations}
According to the Banach fixed-point theorem \cite{agarwal2001fixed}, with a large enough kernel bandwidth $\tau$ and a good initialization $\hbx^{(0)}$, the fixed-point iteration converges to the unique solution of \eqref{eq:fixed_point_compact}. Detailed expression of $\xi$, $\Upsilon(\tau)$, and $\Phi(\tau)$ are given as follows.
\begin{equation}
    \label{eq:xi_definition}
    \xi = \frac{
    \sqrt{n}\sum\nolimits^{D}_{i=1}|\mathbf{z}^\ast[i]|\cdot\|\bF^\ast[i]\|_1
    }
    {
    \lambda_{\text{min}}\big(\sum\nolimits^{D}_{i=1}\sigma_i(\bF^\ast)\big)
    },
\end{equation}

\begin{subequations}
    \label{eq:tau_limits}
    \begin{align}
        \label{eq:tau_limits1}
        \Upsilon(\tau) &= \frac{
        \sqrt{n}\sum\nolimits^{D}_{i=1}|\mathbf{z}^\ast[i]|\cdot\|\bF^\ast[i]\|_1
        }
        {
        \lambda_{\text{min}}\big(\sum\nolimits^{D}_{i=1}\tilde{L}_\tau(\bar{e}_i(\bF^\ast))
        \sigma_i(\bF^\ast)\big)
        }, \\
        \label{eq:tau_limits2}
        \Phi(\tau; \rho) &= \frac{
        \splitdfrac{
        4\sqrt{n}\sum\nolimits^{D}_{i=1}\big[\bar{e}_i(\bF^\ast)\|\bF^\ast[i]\|_1
        }
        {\;
        (\rho \|\sigma_i(\bF^\ast)\|_1 + \|\mathbf{z}^\ast[i]\bF^\ast[i]\|)\big]
        } 
        }
        {
        \tau^2\lambda_{\text{min}}\big(\sum\nolimits^{D}_{i=1}\tilde{L}_\tau(\bar{e}_i(\bF^\ast))
        \sigma_i(\bF^\ast)\big)
        },
    \end{align}
\end{subequations}
where $\sigma_i(\bF^\ast) := \bF^\ast[i]{^\top}\bF^\ast[i]$, $\bar{e}_i(\bF^\ast) := |\mathbf{z}^\ast[i]|+\rho\|\bF^\ast[i]\|_1$, and $\tilde{L}_\tau(x) := L^2_{\tau}(x)/(1+x^2)$.
\end{theorem}
\begin{proof}
\guido{
The proof follows the same procedure as that of the filters using the Gaussian kernel \cite{chen2017maximum}.}
\end{proof}

%% file: content_file/4.4-Complexity.tex
\subsection{Computational Complexity}
For robotic applications, the filtering algorithm should have good real-time performance. The MLVC-EKF is more computationally expensive due to the various operations involved in the fixed point iteration. To shed light on the applicability of the MLVC-EKF, we will analyze its computational complexity and evaluate how it scales with the number of agents \chang{as well as the number of fixed-point iterations}. 

The computational complexity will be analyzed in terms of the number of floating-point operations. Specifically, we consider addition (subtraction), multiplication(division), trigonometric operation, matrix inversion, etc. The specific counting rule for the mathematical operations is summarized in Table \ref{tb:counting rule}. \guido{Involved equations that are taken into account are \eqref{eq:ekf_prediction}-\eqref{eq:basic-kalman_gain}, \eqref{eq:error_augmentation}-\eqref{eq:error_aug_modified}, and \eqref{eq:kernel_matrix}-\eqref{eq:kernel_wighted_variance}.}
\begin{table}[htbp]
\renewcommand\arraystretch{1.2}\small
  \centering
  \caption{Counting Rule of Mathematical Operations}
  \label{tb:counting rule}
  \begin{tabular}{l|c}
    \toprule
        Math Operation & Counting \\
    \midrule
    \midrule
        \chang{$M^{-1}, \CH(M) \; (M\in \mathbb{R}^{n\times n})$} & $\mathcal{O}(n^3)$ \\
        \chang{$M_1 \cdot M_2 (M_1\in \mathbb{R}^{m\times n}, M_2\in \mathbb{R}^{n\times p})$} & \chang{$\mathcal{O}(mnp)$} \\
        $M_1 \pm M_2\; (M_1, M_2\in \mathbb{R}^{m\times n})$ & $\mathcal{O}(mn)$ \\
        $\sin(a), \cos(a), \ln(a) \; (a \in \mathbb{R})$ & $\mathcal{O}(1)$ \\
    \bottomrule
 \end{tabular}
\end{table}
\chang{The number of floating point operations is associated with the number of neighbors $N_i$, the number of fixed-point iterations $T(k)$, and the number of indirect measurements $p_{\text{id}}$. We consider the sufficient worst case in which full-observability is ensured such that $p_{\text{id}} = 2N_i$ and choose $T_m$ as the averaged number of fixed-point iterations. Thus, the final results will be only dependent on $N_i$ and $T_m$, and the details are presented in Table \ref{tb:complexity}, where we separately compute the complexity for each of the terms $\mathcal{O}(N_i)$, $\mathcal{O}(N^2_i)$ and $\mathcal{O}(N^3_i)$.}
\begin{table}[htbp]
\renewcommand\arraystretch{1.05}\small
  \centering
  \caption{\chang{Computational Complexity}}
  \label{tb:complexity}
  \begin{tabular}{c|c|c|c}
    \toprule
      CRL Scheme &  $\mathcal{O}(N_i)$ & $\mathcal{O}(N^2_i)$ & $\mathcal{O}(N^3_i)$\\
    \midrule
      \multicolumn{4}{c}{EKF Estimation} \\
    \midrule
      nCRL & $888$ & $0$ & $0$ \\
      hCRL & $187$ & $289$ & $335$ \\
      fCRL & $224$ & $309$ & $597$ \\
    \midrule
        \multicolumn{4}{c}{MLVC-EKF Estimation} \\
    \midrule
      nCRL& $975+250T_m$ & $0$ & $0$ \\
      hCRL & $189+56T_m$ & $309+5T_m$ & $415+195T_m$ \\
      fCRL & $223+72T_m$ & $325+21T_m$ & $569+401T_m$ \\
    \bottomrule
 \end{tabular}
\end{table}
\chang{
From the results in Table \ref{tb:complexity}, the complexity can be high when the number of neighboring agents $N_i$ is big, especially for hCRL and fCRL schemes where $\mathcal{O}(N^2_i)$ and $\mathcal{O}(N^3_i)$ terms are present due to the matrix multiplication and inversion operations on high dimensional matrices. Another factor that greatly influences the complexity is the average number of fixed-point iterations $T_m$ which scales badly regarding $\mathcal{O}(N^3_i)$. In real experiments, it might be necessary to restrict the maximum allowed number of iterations when using kernel-induced EKF.}

%% file: content_file/5-simulation_results.tex
\section{Simulation Results \& Analysis}
\label{sec:5-simulation}
In this section, the simulation results will be presented to demonstrate the advantages of using the fCRL scheme combined with the MLVC-EKF. We will first describe the \sven{simulation environment and setup}. Next, we compare different CRL methods, in which we investigate all three CRL models with the EKF and the MLVC-EKF. Finally, in the fCRL framework, we compare the kernel-induced Kalman filter with different kernels to show the advantages of the Logarithmic-Versoria kernel.

\input{content_file/5.1-modeling-setting.tex}

\input{content_file/5.2-crl-comparison.tex}

\input{content_file/5.3-kernel-comparison.tex}

%% file: content_file/5.1-modeling-setting.tex
\subsection{\sven{Modeling Setup \& Parameters}}
In the following description, the unit of time, position (distance) and angle are unanimously denoted as \textit{second} $\tunit$, \textit{meter} $\munit$ and \textit{radius} $\aunit$, respectively. We simulate a group of $5$ agents ($N = 5$) with each of them flying the same type of nominal trajectory. The position trajectory is given as
\begin{equation}
    \label{eq:nominal_traj}
    \left \{
    \begin{aligned}
        x(t) &= x_c + R\cos(2\pi ft + \vartheta) \\
        y(t) &= y_c + R\sin(2\pi ft + \vartheta) \\
        z(t) &= z_c + R_z\sin(2\pi f_zt) \\
    \end{aligned}
    \right. ,
\end{equation}
where \chang{$t \in [0 \mathrm{s}, 30 \mathrm{s}]$} is the time, and $x(t)$, $y(t)$, $z(t)$ are all global positions in a preset common reference frame. \guido{The trajectories are designed such that the observability is stimulated infinitely often}. As for the heading, each agent has \guido{a different} initial heading $\psi_0$, and it performs \textit{intermittent} heading turning within the simulation horizon. Specifically, an agent changes its heading of a fixed angle $\psi_\text{t}$ during time interval $[t_i, t_i + 2]$ at a constant heading rate. The parameters describing the nominal trajectory (i.e., both the positions and the heading) for each of the agents are summarized in Table \ref{tb:nominal_traj}.
\begin{table}[htbp]
\renewcommand\arraystretch{1.05}\small
  \centering
  \caption{\chang{Nominal Trajectory Parameters}}
  \label{tb:nominal_traj}
  \begin{tabular}{c|c|c|c|c|c}
    \toprule
        Agent & $1$ & $2$ & $3$ & $4$ & $5$ \\
    \midrule
    \midrule
    \multicolumn{6}{c}{Position Parameters} \\
    \midrule
        $x_c \munit$ & $0$ & $2$ & $-2$ & $-2$ & $2$ \\
        $y_c \munit$ & $0$ & $2$ & $2$ & $-2$ & $-2$ \\
        $z_c \munit$ & $7$ & $8$ & $9$ & $6$ & $5$ \\
        $R \munit$ & $1$ & $1.2$ & $0.8$ & $1.3$ & $0.7$ \\
        $R_z \munit$ & $4$ & $4.5$ & $6$ & $3.5$ & $2$ \\
        $f \funit$ & $0.3$ & $0.4$ & $0.2$ & $0.5$ & $0.1$ \\
        $f_z \funit$ & $0.2$ & $0.4$ & $0.3$ & $0.35$ & $0.25$ \\
        $\vartheta \aunit$ & $0$ & $\pi/4$ & $4\pi/3$ & $-4\pi/3$ & $-\pi/4$ \\
    \midrule
    \multicolumn{6}{c}{Heading Parameters} \\
    \midrule
        $\psi_0 \aunit$ & $0$ & $2\pi/5$ & $3\pi/5$ & $4\pi/5$ & $\pi/5$ \\
        $\psi_\text{t} \aunit$ & $\pi/6$ & $\pi/6$ & $-\pi/6$ & $-\pi/6$ & $\pi/6$ \\
        $t_i \tunit$ & $3;10;20$ & $6;12;15$ & $4;8;11$ & $5;9;12$ & $7;11;25$ \\
    \bottomrule
 \end{tabular}
\end{table}
The nominal trajectory described above still differs from the real trajectory due to actuator noise. In this work, we assume the actuator noise $\Delta \uu_i = [n_{\psi}, n_{v,x}, n_{v,y}, n_{v,z}]^\top$ obeys a zero-mean Gaussian distribution with a \textit{constant} covariance matrix $\mathbf{Q}_{\uu} = \diag\{\sigma^2_{\psi}, \sigma^2_{v}, \sigma^2_{v}, \sigma^2_{v}\}$, where $\sigma_{\psi} = 0.4\avunit$ and $\sigma_{v} = 0.25 \vunit$. Based on the nominal trajectory, we can easily compute the nominal velocities as
\begin{equation}
    \label{eq:nominal_velocity}
    \left \{
    \begin{aligned}
        v_x(t) &= -2\pi f R\sin(2\pi ft + \vartheta) \\
        v_y(t) &= 2\pi f R\cos(2\pi ft + \vartheta) \\
        v_y(t) &= 2\pi f_z R_z\cos(2\pi f_zt) \\
    \end{aligned}
    \right. ,
\end{equation}
\begin{equation}
    \label{eq:nominal_angularv}
    \dot{\psi}(t) = \left \{
    \begin{aligned}
        & 0.5\psi_t,  t \in [t_i, t_i + 2] \\
        & 0, \quad \quad \text{otherwise} 
    \end{aligned}
    \right. .
\end{equation}
The real trajectory of each agent will be generated, given the initial positions ($x_c, y_c, z_c$) and heading ($\psi_0$), using the noisy velocity (i.e., the nominal velocity plus the actuator noise).

The PDF of the basic UWB noise follows \eqref{eq:noise_pdf_sven}, and we adopt the validated parameters from \cite{pfeiffer2021computationally}. As for the delay effects, the PDF of $\nu_d$ depends on both $d_{jl}$ and $\bar{r} = \bar{\eta}\bar{v}$, and it is measurement dependent and thus time-varying. In general, a smaller $d_{jl}$ results in a larger variance, thus making $\nu_d$ more tricky to handle. For modeling simplicity, we take $d_{jl} = 3\bar{r}$ as the worst case to make the noise distribution independent of time, and the corresponding distribution will thus only depends on $\bar{\eta}$ and $\bar{v}$. The parameters governing the noise distributions are outlined in Table \ref{tb:noise_parameters}. 
\begin{table}[htbp]
\renewcommand\arraystretch{1.05}\small
  \centering
  \caption{Measurement Noise Parameters}
  \label{tb:noise_parameters}
  \begin{tabular}{c|c|c|c}
    \toprule
        \multicolumn{4}{c}{UWB Noise $\nu$} \\
    \midrule
        $\sht$ & $0.2$ & $\mu \munit$ & $0.1$ \\
        $\lambda$ & $3.5$ & $k$ & $2$ \\
        $\sigma \munit$ & $0.1$ & & \\
    \midrule
        \multicolumn{4}{c}{Delay Effects Noise $\nu_d$} \\
    \midrule
        $\bar{\eta} \tunit$ & $0.01$ & $\bar{v} \vunit$ & $15$ \\
    \bottomrule
 \end{tabular}
\end{table}

In the simulation, the filter run at the discrete time $k\Delta t$ with the sampling interval $\Delta t = 0.01 \tunit$. When doing state estimation, the filter receives the body velocities and heading rate derived from the nominal input for prediction and then uses the noisy distance measurements for state correction. In the following, we will elaborate on the initialization settings and the noise covariance settings of the filter.

First, the filter does not know the true initial state (i.e., the true relative position and relative heading). Given the true initial \textit{global} state tuple $\{\psi_0, x_c\lsjia R\cos(\varphi), y_c\lsjia R\sin(\varphi), z_c\}$ of all the $5$ agents, the true initial state $\x_{ij}(0)$ for any pair $(i,j)$ can be computed, based on which the true state $\x_i(0)$ for the CRL task of agent $i$ can be built by state augmentation. Then, we manually add initial offset $\Delta \x_{ij}(0) = [\Delta \psi_0, \Delta x_0, \Delta y_0, \Delta z_0]^\top$ to the initial state, where the heading uncertainty is simply $\psi_0 \sim U(-\bar{\psi}_e, \bar{\psi}_e)$ and the position uncertainty is designed as 
\begin{equation}
    \label{eq:position uncertainty}
    \left \{
    \begin{aligned}
        \Delta x_0 &= \bar{r}_e \cos(\varrho)\cos(\varphi) \\
        \Delta y_0 &= \bar{r}_e \cos(\varrho) \sin(\varphi) \\
        \Delta z_0 &= \bar{r}_e \sin(\varrho)
    \end{aligned}
    \right. ,
\end{equation}
where $\varrho \sim U(-\pi/2, \pi/2)$ and $\varphi \sim U(0, 2\pi)$. In other words, the design rule in \eqref{eq:position uncertainty} ensures that the initial position error has a fixed length $\bar{r}_e$ whereas its direction is unknown. Following the state augmentation modeling in Section \ref{sec:4-mainapp}, for an agent $i$ and its neighbors $\mathcal{N}_i$, we first build $\Delta \x_i(0) = \textstyle\vect^{N_i}_{\alpha=1}\{\Delta \x_{ij_\alpha}(0)\}$ where $\Delta \x_{ij_\alpha}(0)$ are repeatedly generated using the same \textit{uncertainty level} pair $(\bar{\psi}_e, \bar{r}_e)$, then the initial state of the filter is given by $\hat{\x}^+(0) = \x_i(0) + \Delta \x_i(0)$. Correspondingly, the initial covariance is set as $\bP^+(0) = \mathds{I}_{N_i} \otimes \bP_e$, where $\bP_e = \diag\{\bar{\psi}_e^2/3, \bar{r}_e^2/4, \bar{r}_e^2/4, \bar{r}_e^2/2\}$. 

Apart from the initialization, the filters should have some prior knowledge of the noise. \chang{The velocity input uncertainty is determined by the onboard hardware (i.e., IMU, gyroscope) of the MAV itself, which means the actuator noise statistics can be obtained \textit{offline beforehand}. Therefore, we assume that the filter has access to the true actuator covariance matrix $\mathbf{Q}_{\uu}$.} For nCRL scheme where pairwise localization is the foundation, we set $\Q_k = \diag\{\Q_{\uu}, \Q_{\uu}\}$ (cf. step 22 of Algorithm \ref{algorithm2}), whereas for fCRL and hCRL schemes, we have $\Q_k = \mathds{I}_{N_i+1} \otimes \Q_{\uu}$ (cf. step 14 of Algorithm \ref{algorithm2}). 

\chang{The UWB noise ideally is Gaussian (i.e., $\sht = 0$ in \eqref{eq:noise_pdf_sven}) without accounting for practical issues (e.g., NLoS, multi-pathing, transmission delay), and this ideal UWB noise information which is only related to the UWB device itself is also available \textit{beforehand}. The remaining tricky part is the heavy-tailed component of the UWB noise which intrinsically is environment-dependent as well as task-dependent. As a consequence, the overall UWB noise information cannot be accurately obtained in general, and how to assign a proper measurement covariance matrix $\bR_k$ to the Kalman-type filters is difficult and may require additional adaptive or online learning techniques which significantly increase computational complexity. In this regard, we propose the following two initialization strategies for $\bR_k$.
\begin{itemize}
    \item \textbf{Smart}: The filter has access to the true heavy-tailed noise statistics, and $\bR_k$ is \textit{well} initialized to balance the heavy-tailed component.
    \item \textbf{Inattentive}: The filter only has information about the basic Gaussian statistics of the UWB noise, and $\bR_k$ is \textit{badly} initialized.
\end{itemize}}
Based on the noise parameters given in Table \ref{tb:noise_parameters}, we can numerically obtain $\sigma^2_{\nu} \approx 0.08$ and $\sigma^2_{\nu_d} \approx 0.01$. The details of the measurement covariance matrix $\mathbf{R}_k$ for both of the above strategies are summarized in Table \ref{tb:meas_noise_variance}.
\begin{table}[htbp]
\renewcommand\arraystretch{1.2}\small
  \centering
  \caption{Measurement Covariance Matrix $\mathbf{R}_k$}
  \label{tb:meas_noise_variance}
  \begin{tabular}{c|c|c}
    \toprule
         & Smart & Inattentive \\
    \midrule
    \midrule
        fCRL & $\diag\{0.08\mathds{I}_{N_i}, 0.09\mathds{I}_{p_{\text{id}}}\}$ & $0.01\mathds{I}_{(N_i+p_{\text{id}})}$ \\
        hCRL & $0.08\mathds{I}_{N_i}$ & $0.01\mathds{I}_{N_i}$ \\
        nCRL & $0.08$ & $0.01$ \\
    \bottomrule
 \end{tabular}
\end{table}

In our simulation, we assume that for any agent $i \in \{1,2,3,4,5\}$, its neighbor set is $\mathcal{N}_i = \{1,2,3,4,5\}\setminus\{i\}$ (i.e, all 5 agents are closed neighbors to each other). Given the above condition, for fCRL scheme, the number of indirect measurements is $p_{\text{id}} = \tbinom{4}{2} = 6$. We select agent $1$ to be the interested agent that is going to localize its neighbors.

%% file: content_file/5.2-crl-comparison.tex
\subsection{Comparison of Different CRL Methods}
For each of the 3 localization schemes (i.e., fCRL, hCRL, nCRL), it can be combined with either the MLVC-EKF or the EKF. Thus, we have $2 \times 3 = 6$ different comparable CRL methods in total. For a given scheme xCRL, if it is combined with MLVC-EKF, the corresponding method will be abbreviated as xCRL(MLVC) (e.g., fCRL(MLVC) stands for the method based on the fCRL scheme using the MLVC-EKF estimation algorithm), otherwise, it will be simply denoted as xCRL if its embedded filter is the EKF.

To compare the different CRL methods, we run a Monte Carlo simulation in which we vary both the initialization setting and the noise sequences. \chang{We first set $6$ uncertainty level pairs $\{(\bar{\psi}^q_e, \bar{r}^q_e)\}$ with $(\bar{\psi}^q_e, \bar{r}^q_e) = (q\pi/18, q/\chang{2}), (q = 1,2,\cdots,6)$. The unit of $\bar{\psi}^q_e$ and $\bar{r}^q_e$ is [rad] and [m], respectively.} Then, for a specific uncertainty level pair, we run $20$ trials where each trial has its unique specific initialization and it is simulated with randomly generated noise sequences (for both actuator noise and measurement noise). Therefore, for a given CRL method, we run \chang{$6 \times 20 = 120$} trials in total.

For each trial, we will compute both the averaged heading error and the averaged position error, \chang{and the mathematical expression of the four considered error metrics are given as}
\chang{
\begin{subequations}
\label{eq:error_statistical_criterion}
    \begin{align}
        \text{TR-ER}_{\psi} &= \frac{1}{4N_s} \sum^{4}_{\alpha = 1}\sum^{k_0}_{\chang{k = 1}}|\psi_{1j_\alpha}(k) - \hat{\psi}_{1j_\alpha}(k)|, \\
        \text{SS-ER}_{\psi} &= \frac{1}{4N_s} \sum^{4}_{\alpha = 1}\sum^{N_s}_{\chang{k = k_0+1}}|\psi_{1j_\alpha}(k) - \hat{\psi}_{1j_\alpha}(k)|, \\
        \text{TR-ER}_{\p} &= \frac{1}{4N_s} \sum^{4}_{\alpha = 1}\sum^{k_0}_{\chang{k = 1}}\|\p_{1j_\alpha}(k) - \hat{\p}_{1j_\alpha}(k)\|, \\
        \text{SS-ER}_{\p} &= \frac{1}{4N_s} \sum^{4}_{\alpha = 1}\sum^{N_s}_{\chang{k = k_0+1}}\|\p_{1j_\alpha}(k) - \hat{\p}_{1j_\alpha}(k)\|,
    \end{align}
\end{subequations}
where $N_s = 30/T_s = 3\times 10^{3}$ is the total number of discretized time stamps, $k_0 = 10^3$ is the separation time stamp which separates the entire simulation horizon into the first $0\mathrm{s} -10\mathrm{s}$ \textit{transient} interval and the last $10\mathrm{s}-30\mathrm{s}$ \textit{steady-state} interval. Following the previous convention, $j_\alpha \in \mathcal{N}_1 = \{2,3,4,5\}$ denotes the neighbors of agent $1$, the variables without hat (i.e., $\psi_{1j_\alpha}(k)$ and $\p_{1j_\alpha}(k)$) are \textit{ground truth} derived from the real trajectory and evaluated at $kT_s$, and the variables with hat (i.e., $\hat{\psi}_{1j_\alpha}(k)$ and $\hat{\p}_{1j_\alpha}(k)$) are the estimated state using a specific CRL method.}

\begin{figure} 
    \centering
  \subfloat[Heading Error]{%
       \includegraphics[width=\linewidth]{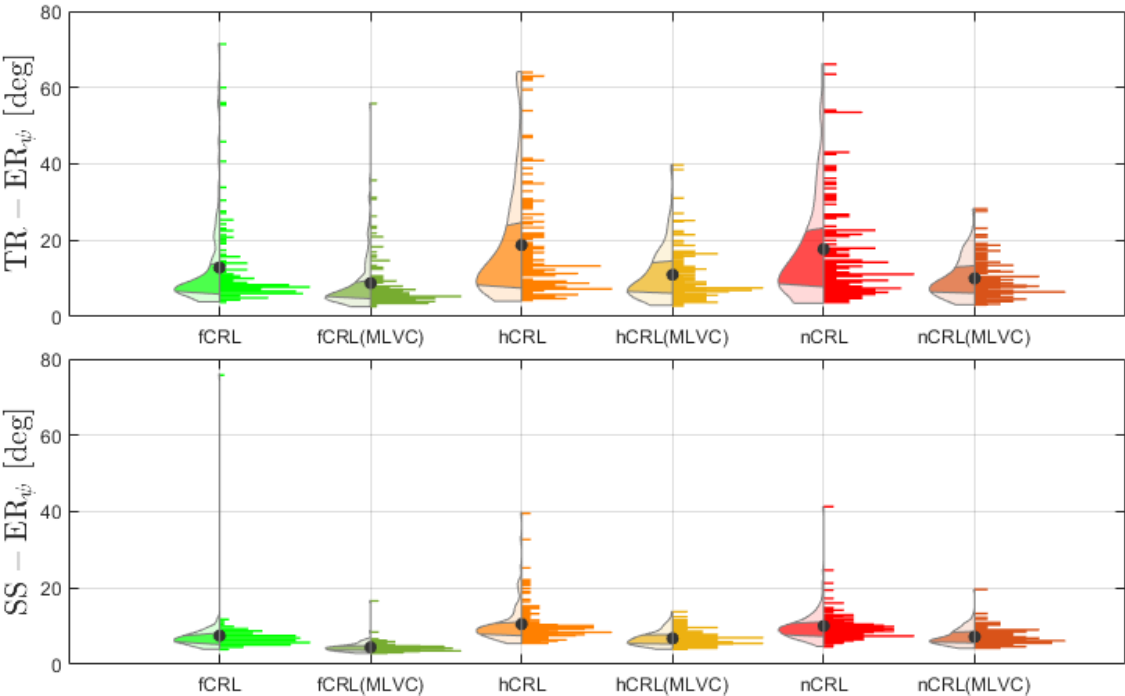}
       \label{fig:violin_smart(heading)}}
    \\
  \subfloat[Position Error]{%
        \includegraphics[width=\linewidth]{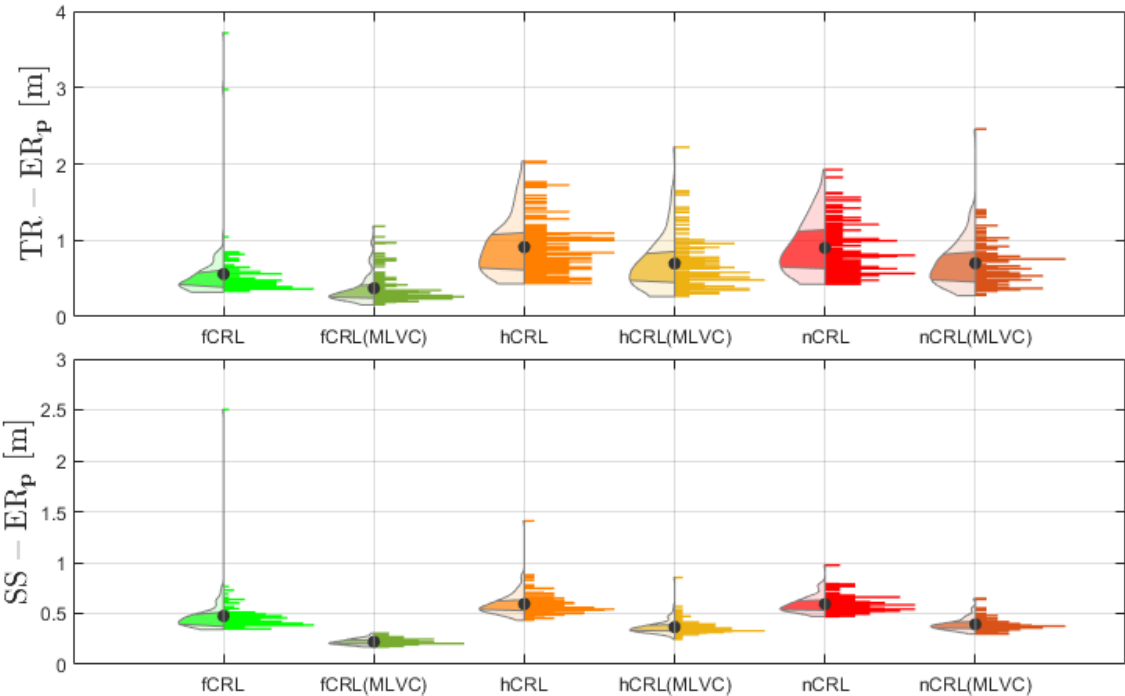}
        \label{fig:violin_smart(position)}}
  \caption{Violin Plots of the Monte Carlo simulation for all 6 CRL methods with all filters in \textbf{smart} mode, (a) Averaged Heading Error (b) Averaged Position Error. For each subfigure (a) and (b), from left to right: 1) fCRL, 2) fCRL(MLVC), 3) hCRL, 4) hCRL(MLVC), 5) nCRL,  6) nCRL(MLVC). \chang{For each violin plot, the averaged error metric with respect to the trials is marked as a black dot, the left part is the approximated PDF of the data with the two boundaries of the shaded area representing the $25\%$ and $75\%$ quantile, and the right part is the histogram.}}
  \label{fig:violin_smart} 
\end{figure}

\chang{First, we carried out the Monte Carlo simulation where the embedded filter of each of the CRL methods is in \textit{smart} mode.} For a given CRL method, the corresponding performance evaluation quantities $\text{TR-ER}_\psi$, $\text{SS-ER}_\psi$, $\text{TR-ER}_{\p}$ and $\text{SS-ER}_{\p}$ \chang{were} computed for all the \chang{$120$} trials. \chang{The results are shown through the violin statistical plots in Fig. \ref{fig:violin_smart}, where Fig. \ref{fig:violin_smart(heading)} and Fig. \ref{fig:violin_smart(position)} summarize the $\text{TR-ER}_\psi$ ($\text{SS-ER}_\psi$) data and $\text{TR-ER}_{\p}$ ($\text{SS-ER}_{\p}$) data, respectively.}

\begin{figure*}
    \centering
    \includegraphics[width=\textwidth]{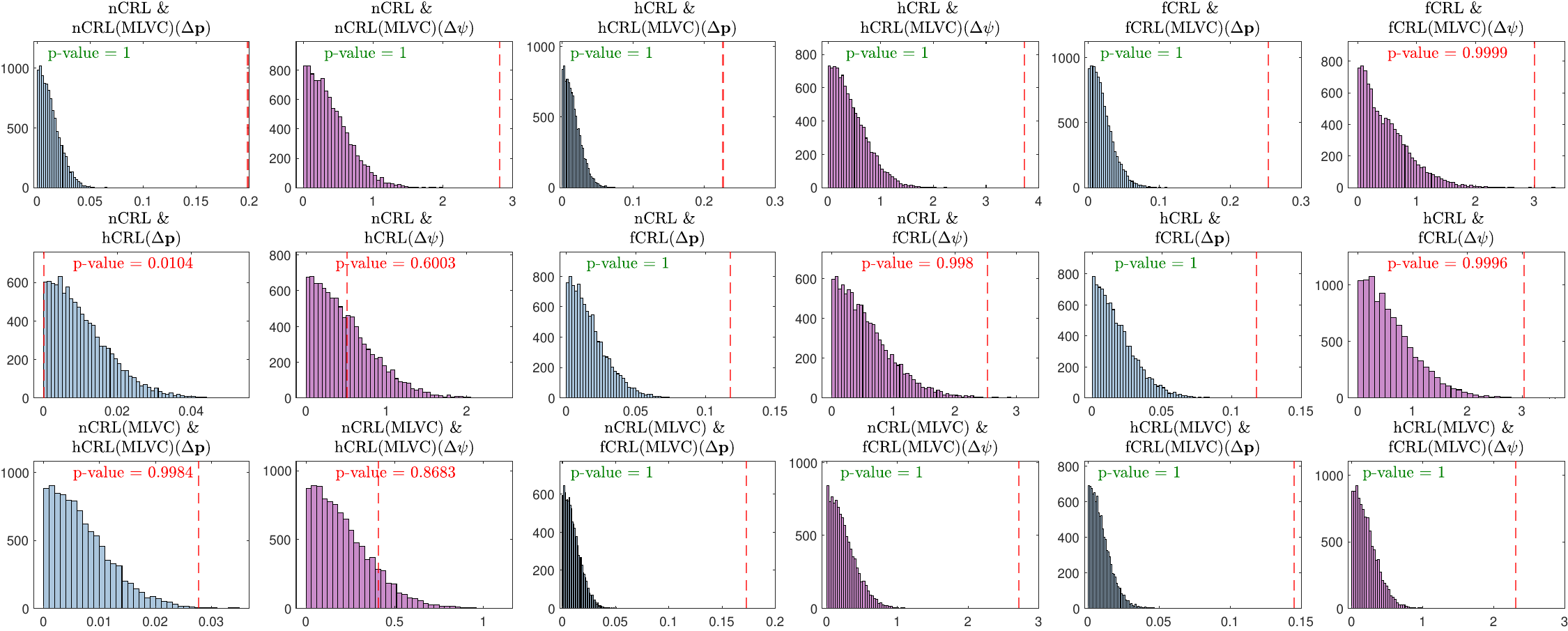}
    \caption{\chang{Bootstrapping statistical analysis for evaluating the steady-state performance of all comparable CRL methods with \textbf{smart mode} filters}}
    \label{fig:boot_smart}
\end{figure*}

\chang{
According to \ref{fig:violin_smart}, it can be clearly observed that, for both transient and steady-state performance, the fCRL gives the best estimation results among the 3 CRL schemes for having lower average position and heading error in general. In addition, for all three CRL schemes, the MLVC-EKF exhibits strong power to reject outliers, especially for the fCRL scheme where more measurements are present. To demonstrate the improvements of using fCRL combined with the MLVC-EKF, we apply the bootstrapping technique to further compare the results between various comparable CRL methods. Taking $\text{SS-ER}_\p$ as an example, the basic steps of the bootstrapping analysis are outlined below, and a similar procedure applies also to $\text{TR-ER}_\p$, $\text{TR-ER}_\psi$, and $\text{SS-ER}_\psi$.
\begin{enumerate}
    \item Fuse the $\text{SS-ER}_\p$ data of two CRL methods (i.e., denoted as $\mathcal{A}^\ast_{\p}$ and $\mathcal{B}^\ast_{\p}$, respectively) to form a hybrid data set of size \guido{$120\times 2 = 240$}.
    \item From the hybrid data set, build data set $\mathcal{A}_{\p}$ and $\mathcal{B}_{\p}$, both being of size $240$, by resampling with replacement.
    \item Compute the average of $\mathcal{A}_{\p}$ and $\mathcal{B}_{\p}$, respectively. The results $\bar{A}_{\p}$ and $\bar{B}_{\p}$ are further used to calculate $\Delta \p = |\bar{A}_{\p} -\bar{B}_{\p}|$.
    \item Repeatedly execute Step 2 and 3 for $N_{\text{bs}}$ times to form the final data set $\{\Delta \p(i)\}^{N_{\text{bs}}}_{i=1}$.
    \item Compute the nominal difference $\Delta \p^{\ast} = |\bar{A}^{\ast}_{\p} -\bar{B}^{\ast}_{\p}|$.
    \item Compute the $p$-value of $\{\Delta \p(i)\}^{N_{\text{bs}}}_{i=1}$ evaluated at $p = \Delta \p^{\ast}$, and the $p$-value represents the \textit{confidence level} of whether the original two data set $\mathcal{A}^\ast_{\p}$ and $\mathcal{B}^\ast_{\p}$ differs.
\end{enumerate}}
In this work, we set \chang{$N_{\text{bs}} = 10^{4}$}, and the bootstrapping results \chang{for evaluating the steady-state performance} are summarized in Fig. \ref{fig:boot_smart}, where the plots for heading error statistics and the position error statistics are, respectively, presented in \textit{blue} and \textit{purple}. For each bootstrapping plot, the nominal average difference is marked with a \textit{red} \chang{vertical} dashed line, and the $p$-value is highlighted in \textit{red} if it is smaller than one (i.e., not $100\%$ confident). \chang{Based on the bootstrapping results in Fig. \ref{fig:boot_smart}, it can be further concluded that, in terms of the steady-state heading estimation performance, there is no significant improvement when using hCRL which simply brings correlation compared to the standard nCRL regardless of the used filter, whereas the position estimation is clearly improved if the MLVC-EKF is applied since the confidence level for comparing nCRL(MLVC) and hCRL(MLVC) is $99.84\%$.}

\begin{figure}
    \centering
  \subfloat[Heading Error]{%
       \includegraphics[width=\linewidth]{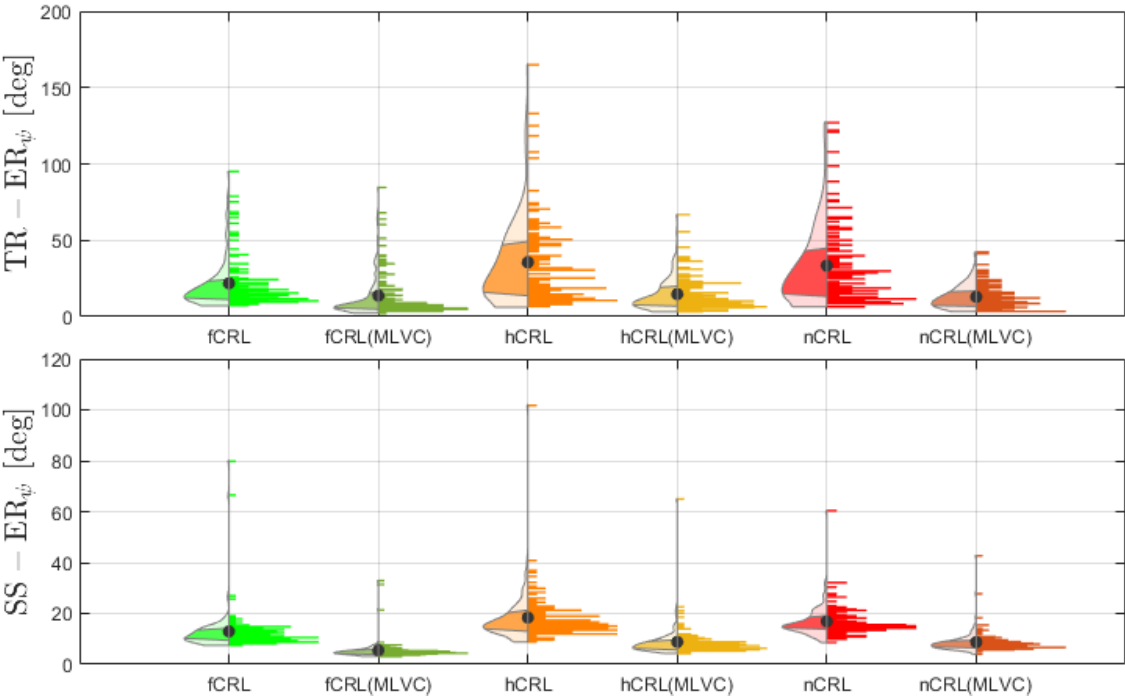}
       \label{fig:violin_inattentive(heading)}}
    \\
  \subfloat[Position Error]{%
        \includegraphics[width=\linewidth]{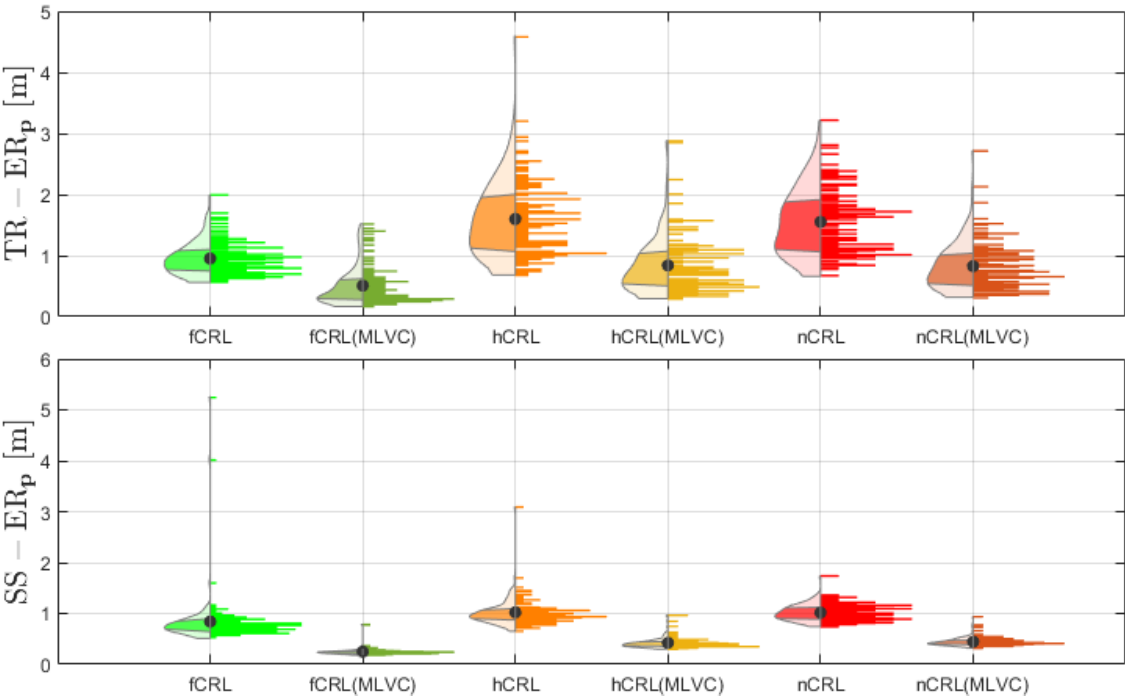}
        \label{fig:violin_inattentive(position)}}
  \caption{Violin plots of the Monte Carlo simulation for all 6 CRL methods with all filters in \textbf{inattentive} mode, (a) Averaged Heading Error (b) Averaged Position Error. For each subfigure (a) and (b), from left to right: 1) fCRL, 2) fCRL(MLVC), 3) hCRL, 4) hCRL(MLVC), 5) nCRL,  6) nCRL(MLVC). \chang{For each violin plot, the averaged error metric with respect to the trials is marked as a black dot, the left part is the approximated PDF of the data with the two boundaries of the shaded area representing the $25\%$ and $75\%$ quantile, and the right part is the histogram.}}
  \label{fig:violin_inattentive} 
\end{figure}

\begin{figure*}
    \centering
    \includegraphics[width=\textwidth]{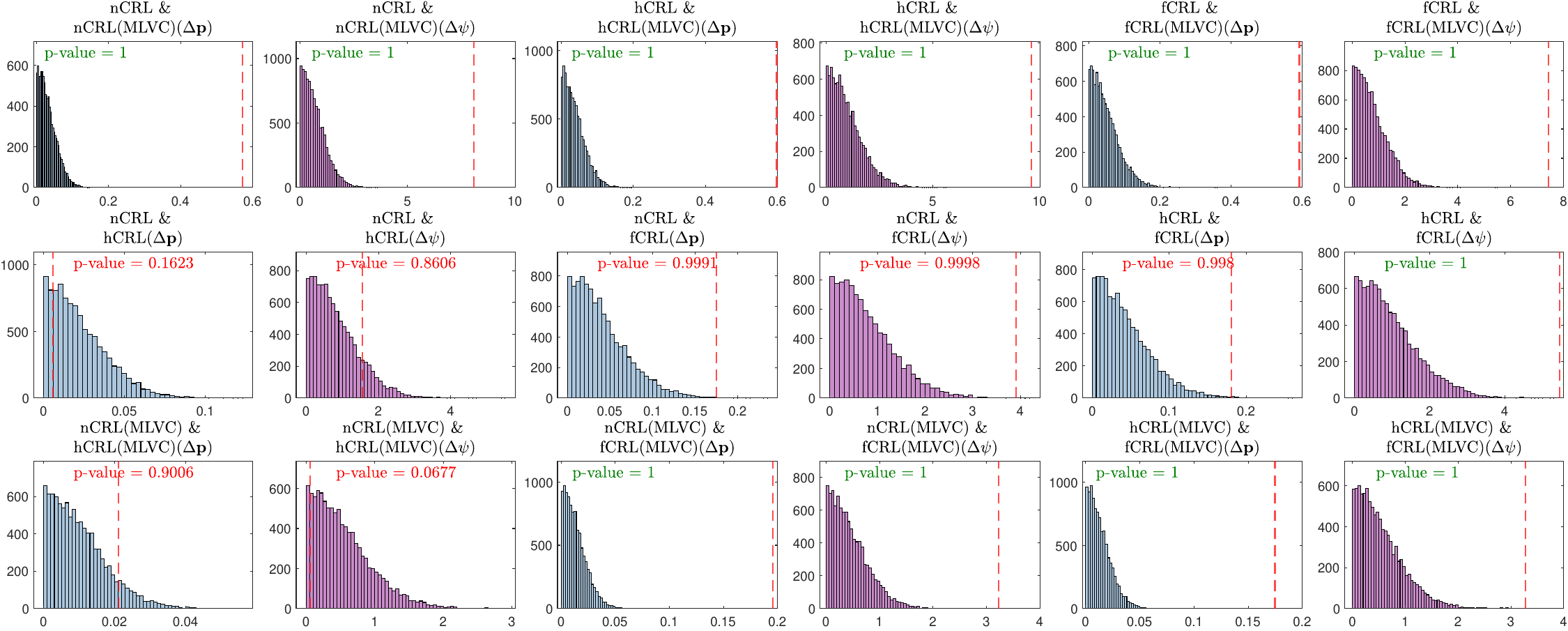}
    \caption{\chang{Bootstrapping statistical analysis for evaluating the steady-state performance of all comparable CRL methods with \textbf{inattentive mode} filters}}
    \label{fig:boot_inattentive}
\end{figure*}

\chang{Next, to see how the CRL methods performs when the filters are initialized with bad matching covariance matrices $\bR_k$, we did the Monte Carlo simulation for all 6 CRL methods with \textit{inattentive mode} filters.} The violin plots are given in Fig. \ref{fig:violin_inattentive}, and the bootstrapping plots are given in Fig. \ref{fig:boot_inattentive}. Based on the results in Fig. \ref{fig:violin_inattentive} and Fig. \ref{fig:boot_inattentive}, we can draw \chang{similar conclusions} that i) the fCRL scheme outperforms the hCRL/nCRL scheme in general, ii) the MLVC-EKF outperforms the EKF regardless of the underlying CRL scheme, \sven{iii) the hCRL scheme does not provide a statistically significant advantage over the nCRL scheme.}

\chang{However, when comparing the magnitude of each of the error metrics presented in Fig. \ref{fig:violin_smart} and Fig. \ref{fig:violin_inattentive}, we can observe that the performance of the inattentive mode filter is worse than that of its smart mode counterpart for all six CRL methods. To highlight the difference between the filters of the smart mode and the inattentive mode, we summarize the average error metrics $\overline{\text{TR-ER}}_\psi$ $\overline{\text{SS-ER}}_\psi$, $\overline{\text{TR-ER}}_{\p}$, and $\overline{\text{TR-ER}}_{\p}$ with respect to all $120$ trials in Table \ref{tb:averagedTRER} and Table \ref{tb:averagedSSER} for transient and steady-state performance, respectively. In both Table \ref{tb:averagedTRER} and Table \ref{tb:averagedSSER}, the percentage increase of the error metrics of CRL method using the inattentive mode filter with respect to that of the same method using smart filter is also computed and highlighted in red color, from which it is clear that the MLVC-EKF is more \textit{robust} than the EKF in terms of handling outliers in cases where a good prior knowledge of the measurement noise statistics is unattainable.}
\begin{table}[htbp]
\renewcommand\arraystretch{1.2}\small
  \centering
  \caption{\chang{Transient Performance (Smart(S) \& Inattentive(I))}}
  \label{tb:averagedTRER}
  \begin{tabular}{c|c|c|c}
    \toprule
         & fCRL & hCRL & nCRL \\
    \midrule
    \midrule
        \multicolumn{4}{c}{Heading $\overline{\text{TR-ER}}_\psi$ [deg]}\\
    \midrule
    EKF & \makecell*[c]{(S)$12.8269$ \\ (I)$21.8339$ \\ \hlt{$70.22\%$}} & \makecell*[c]{(S)$18.7618$ \\ (I)$35.1224$\\ \hlt{$87.20\%$}} & \makecell*[c]{(S)$17.6867$ \\ (I)$33.1865$\\ \hlt{$87.64\%$}} \\
    MLVC-EKF & \makecell*[c]{(S)$\textbf{8.8004}$ \\ (I)$\textbf{13.8284}$\\ \hlt{$57.13\%$}} & \makecell*[c]{(S)$10.9913$ \\ (I)$14.7124$\\ \hlt{$33.86\%$}} & \makecell*[c]{(S)$10.0108$ \\ (I)$13.0007$\\ \hlt{$29.87\%$}} \\
    \midrule
        \multicolumn{4}{c}{Position $\overline{\text{TR-ER}}_\p$ [m]}\\
    \midrule
    EKF & \makecell*[c]{(S)$0.5614$ \\ (I)$0.9397$\\ \hlt{$67.39\%$}} & \makecell*[c]{(S)$0.9191$ \\ (I)$1.6198$\\ \hlt{$76.24\%$}} & \makecell*[c]{(S)$0.8925$ \\ (I)$1.5380$\\ \hlt{$72.32\%$}} \\
    MLVC-EKF & \makecell*[c]{(S)$\textbf{0.3719}$ \\ (I)$\textbf{0.5080}$\\ \hlt{$36.60\%$}} & \makecell*[c]{(S)$0.6947$ \\ (I)$0.8420$\\ \hlt{$21.20\%$}} & \makecell*[c]{(S)$0.6991$ \\ (I)$0.8351$\\ \hlt{$19.45\%$}} \\
    \bottomrule
 \end{tabular}
\end{table}
\begin{table}[htbp]
\renewcommand\arraystretch{1.2}\small
  \centering
  \caption{\chang{Steady-state Performance (Smart(S) \& Inattentive(I))}}
  \label{tb:averagedSSER}
  \begin{tabular}{c|c|c|c}
    \toprule
         & fCRL & hCRL & nCRL \\
    \midrule
    \midrule
        \multicolumn{4}{c}{Heading $\overline{\text{SS-ER}}_\psi$ [deg]}\\
    \midrule
    EKF & \makecell*[c]{(S)$7.5088$ \\ (I)$12.8731$\\ \hlt{$71.44\%$}} & \makecell*[c]{(S)$10.5429$ \\ (I)$18.3855$\\ \hlt{$74.39\%$}} & \makecell*[c]{(S)$10.1980$ \\ (I)$16.3452$\\ \hlt{$60.28\%$}} \\
    MLVC-EKF & \makecell*[c]{(S)$\textbf{4.4992}$ \\ (I)$\textbf{5.4510}$\\ \hlt{$21.15\%$}} & \makecell*[c]{(S)$6.6804$ \\ (I)$8.8255$\\ \hlt{$32.11\%$}} & \makecell*[c]{(S)$7.2203$ \\ (I)$8.7587$\\ \hlt{$21.31\%$}} \\
    \midrule
        \multicolumn{4}{c}{Position $\overline{\text{SS-ER}}_\p$ [m]}\\
    \midrule
    EKF & \makecell*[c]{(S)$0.4758$ \\ (I)$0.8487$\\ \hlt{$78.37\%$}} & \makecell*[c]{(S)$0.5913$ \\ (I)$1.0393$\\ \hlt{$75.77\%$}} & \makecell*[c]{(S)$0.5843$ \\ (I)$1.0189$\\ \hlt{$74.38\%$}} \\
    MLVC-EKF & \makecell*[c]{(S)$\textbf{0.2213}$ \\ (I)$\textbf{0.2488}$\\ \hlt{$12.43\%$}} & \makecell*[c]{(S)$0.3663$ \\ (I)$0.4041$\\ \hlt{$10.21\%$}} & \makecell*[c]{(S)$0.3987$ \\ (I)$0.4622$\\ \hlt{$15.93\%$}} \\
    \bottomrule
 \end{tabular}
\end{table}

\chang{To better visualize the estimation performance and to see the advantages of using fCRL with MLVC-EKF in detail, we also present the state trajectory plots and the top-view plots of some representative trials. To avoid dense plots, we only show the results comparing the baseline CRL method (i.e., nCRL with EKF) and the proposed most advanced CRL method (i.e., fCRL with MLVC-EKF). 

For the top view plots, we present the instantaneous shot at $t = 0\mathrm{s}$, $t = 5\mathrm{s}$, $t = 15\mathrm{s}$, and $t = 20\mathrm{s}$, and the trajectory of the ground truth relative position over a short period following each of the shown time instants is also presented. To also show the relative position along the z-axis, the \textit{size of the markers (agents)} is scaled accordingly. Specifically, the size of the black circle (i.e., agent $1$) is the default size, and if another agent is higher (lower) than agent $1$, it has a relatively bigger (smaller) marker. In addition, the relative heading angle can also be directly viewed in this top-view figure by observing the bar-shaped heading indicator attached to each of the markers (agents). Fig. \ref{fig:top_view} and \ref{fig:top_view2} give the top-view plots for CRL methods with smart mode filters and inattentive mode filters, respectively. The state trajectories are given in Fig. \ref{fig:traj_smart} and Fig. \ref{fig:traj_inattentive}, where all the four states (i.e., heading $\psi_{ij}$, $x_{ij}$, $y_{ij}$, and $z_{ij}$) are given separately, and we only show the state relating to the localization for agent $2$ for simplicity.
\begin{figure*}
    \centering
    \includegraphics[width=\textwidth, trim=5cm 0cm 5cm 0cm]{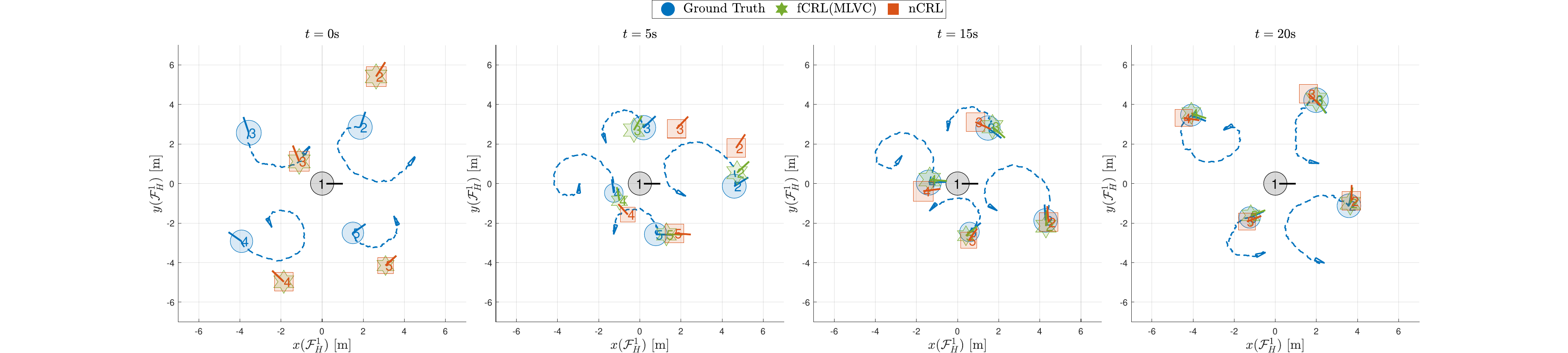}
    \caption{Top view of the localization performance, instantaneous shot at $t = 0\mathrm{s},5\mathrm{s},15\mathrm{s},20\mathrm{s}$ (from left to right). The host agent $1$ is the black circle always located at $(0,0)$, the ground truth agents is the blue circles, the estimated agents using fCRL(MLVC) are the green hexagram, and the estimated agents using nCRL are the red squares. (\textbf{smart} mode filters)}
    \label{fig:top_view}
\end{figure*}
\begin{figure*}
    \centering
    \includegraphics[width=\textwidth, trim=5cm 0cm 5cm 0cm]{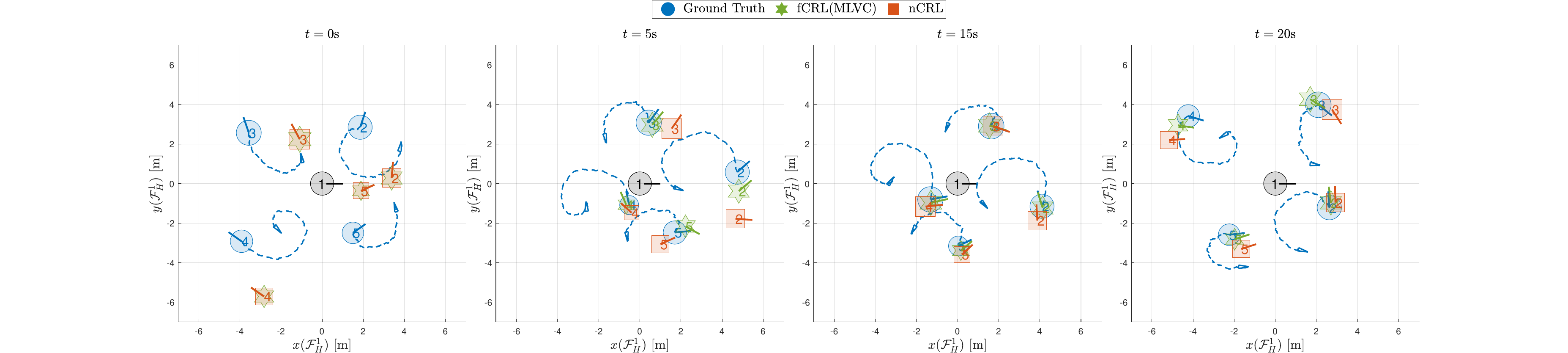}
    \caption{Top view of the localization performance, instantaneous shot at $t = 0\mathrm{s},5\mathrm{s},15\mathrm{s},20\mathrm{s}$ (from left to right). The host agent $1$ is the black circle always located at $(0,0)$, the ground truth agents is the blue circles, the estimated agents using fCRL(MLVC) are the green hexagram, and the estimated agents using nCRL are the red squares. (\textbf{inattentive} mode filters)}
    \label{fig:top_view2}
\end{figure*}
\begin{figure}
    \centering
    \includegraphics[width=\linewidth]{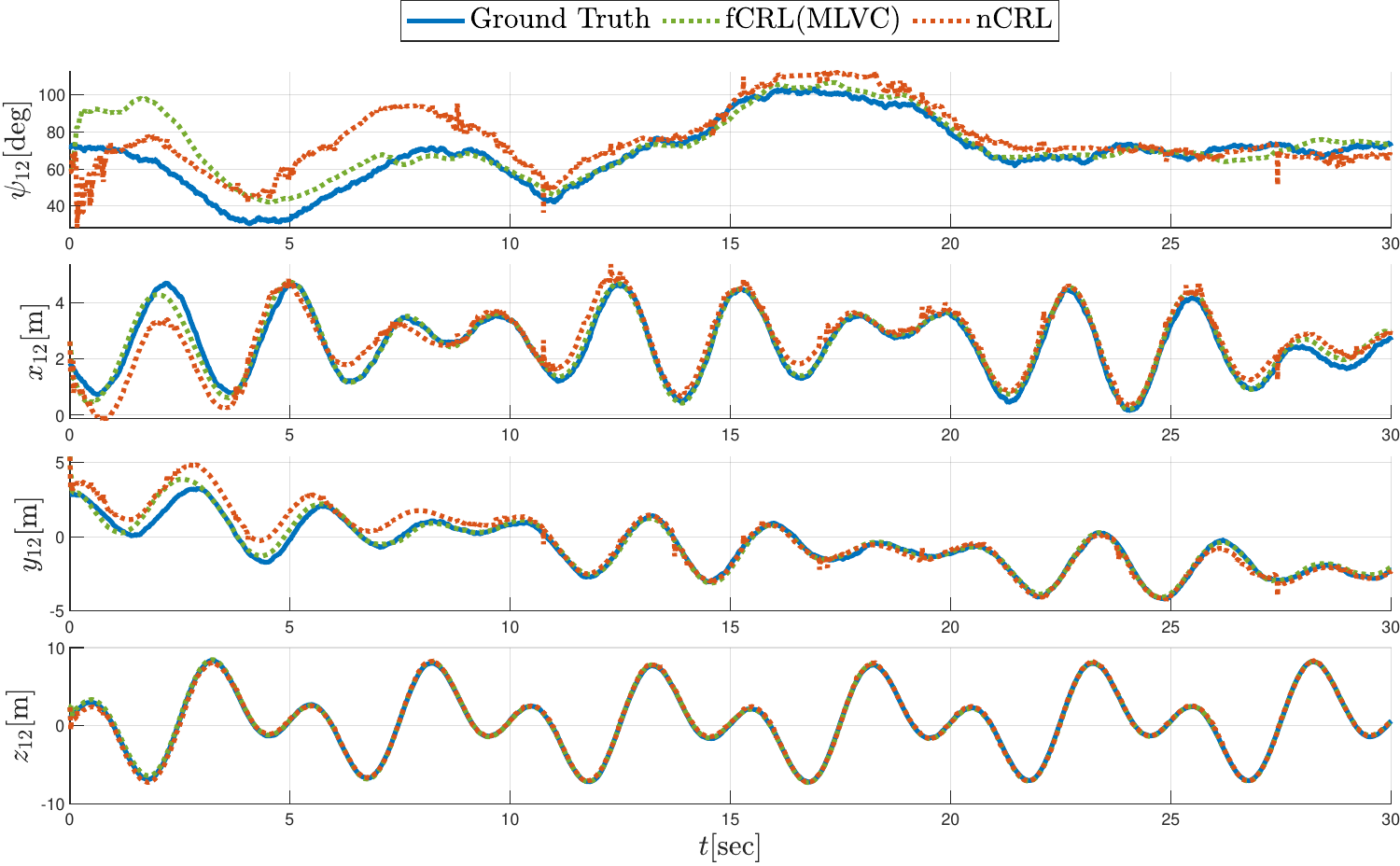}
    \caption{State trajectory of the localization for agent $2$ (smart mode filters)}
    \label{fig:traj_smart}
\end{figure}
\begin{figure}
    \centering
    \includegraphics[width=\linewidth]{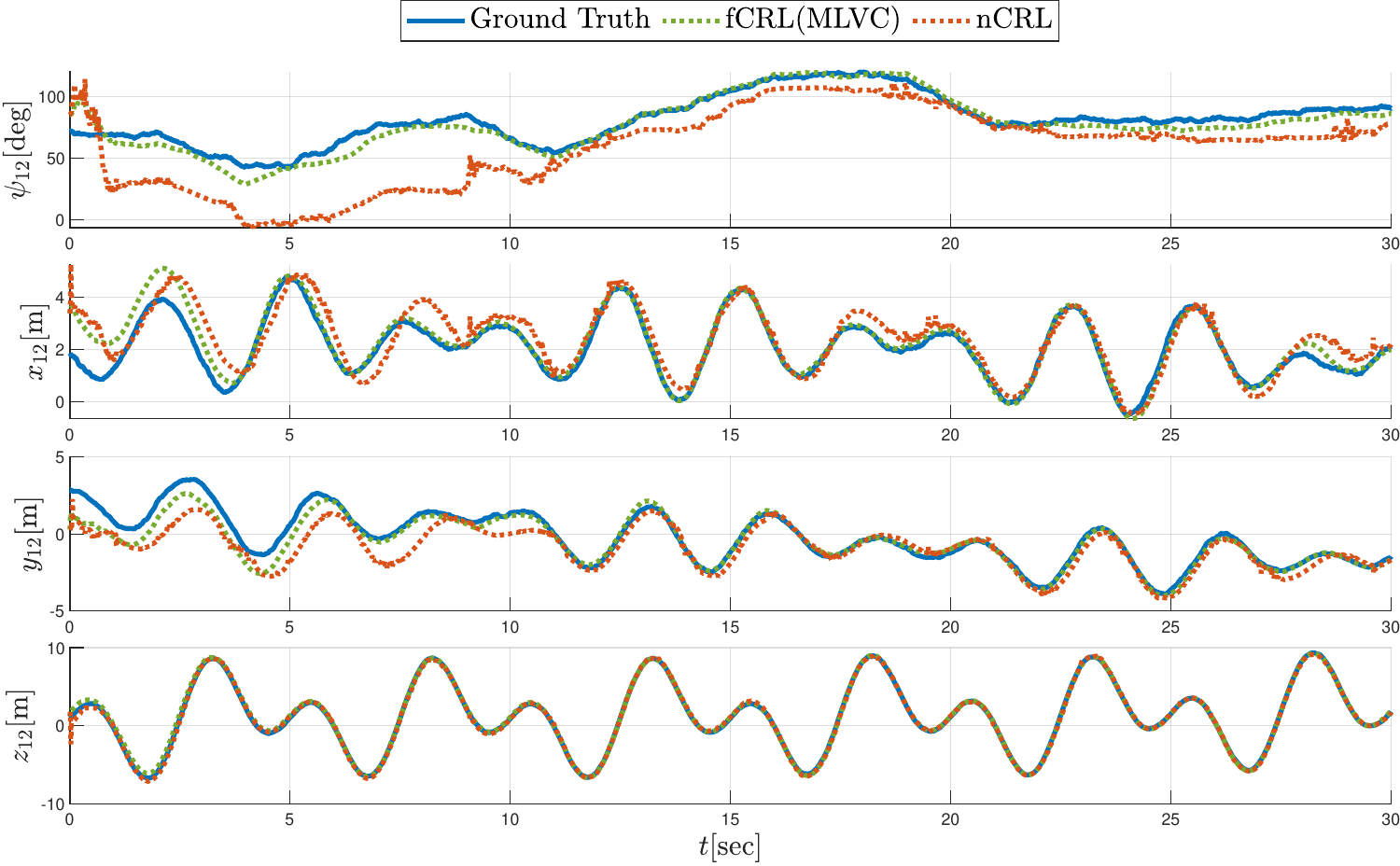}
    \caption{State trajectory of the localization for agent $2$ (inattentive mode filters)}
    \label{fig:traj_inattentive}
\end{figure}

From both the state trajectory and the top-view plots, it can be observed that, compared to the nCRL method, the fCRL(MLVC) method converges faster in the transient interval and also provides a more accurate estimation in the steady-state interval. Moreover, the trajectory of nCRL method exhibits frequent undesirable \textit{sharp peaks} and it also deviates from the ground truth trajectory even after convergence, especially in cases where the inattentive mode filters are used.}

During the simulation, we also recorded the run time of different CRL methods. For a given scheme, the run time of each iteration $k$ during trial $i$ is denoted as $\Delta t(i, k)$, and this time cost is recorded for all time instant and for all $120$ trials. The average time cost $\overline{\Delta t}$ used for evaluation is computed as
\begin{equation}
    \label{eq:timecost}
    \chang{\overline{\Delta t}} = \frac{1}{200N_s}\sum^{200}_{i=1}\sum^{N_s}_{k=1} \Delta t(i, k), \notag
\end{equation}
where $N_s = 30/T_s = 3\times 10^3$, and the results for all 6 CRL methods are given in Table \ref{tb:runtime}. Compared with the baseline nCRL method, though using fCRL/hCRL and \chang{incorporating} MLVC-EKF both increase the computational time. \sven{Specifically, the computational time increases by about $2\sim 3\%$ when bringing more measurements (i.e., from hCRL to fCRL), and it dramatically increases by about $260\%$ when considering the correlation (i.e., from nCRL to hCRL). As for incorporating MLVC-EKF, the computational time increases by about $11\%$, $15\%$ and $57\%$ for fCRL, hCRL and nCRL, respectively.}
\begin{table}[htbp]
\renewcommand\arraystretch{1.2}\small
  \centering
  \caption{Average Run Time (unit: $10^{-1}\mathrm{ms}$)}
  \label{tb:runtime}
  \begin{tabular}{c|c|c|c}
    \toprule
         & fCRL & hCRL & nCRL \\
    \midrule
    \midrule
    EKF & $6.2822$ & $6.0855$ & $2.3901$ \\
    MLVC-EKF & \guido{$7.0136$} & $6.9214$ & $3.7712$ \\
    \bottomrule
 \end{tabular}
\end{table}

%% file: content_file/5.3-kernel-comparison.tex
\subsection{Comparison of different kernels}
\chang{In this section, we are going to compare the performance of the kernel-induced EKF with different kernels. We only consider applying the filters to the fCRL scheme, and we stick to the \textit{smart mode} filters in terms of setting measurement covariance matrix $\bR_k$. We will compare the designed LV kernel (cf. \eqref{eq:ker_definition}), the Versoria kernel, and the Gaussian kernel. The Versora kernel and Gaussian kernel are given as
\begin{subequations}
    \label{eq:otherkernel}
    \begin{align}
        \label{eq:versoria_kernel}
        V_{\tau}(e) &= \frac{\tau}{\tau + e^2}, \\
        \label{eq:gaussian_kernel}
        G_{\tau}(e) &= \exp(-\frac{e^2}{\tau}),
    \end{align}
\end{subequations}
where $\tau$ is the bandwidth of each of the two kernels. \guido{For the Gaussian kernel and Versoria kernel}, the corresponding maximum correntropy filter will degenerate to the standard EKF and loses its ability to handle non-Gaussian heavy-tailed noise if $\tau \to \infty$. \guido{One advantage of the LV kernel is that the corresponding filter never degenerates and the designer can simply choose a large enough bandwidth if he/she would like to avoid the divergence issue of the fixed point iteration without checking the conditions given in \ref{theorem1}.} In general, choosing a smaller bandwidth leads to a more robust filtering performance while also a larger number of iterations required in the fixed point iteration on average \cite{chen2017maximum}. To compare the performance of the filters when the kernel-based measurement update is effective, we set $\tau = 5.0$ for all three kernels which also guarantees the convergence of the fixed-point iteration as stated in Theorem \ref{theorem1}. We follow the Monte Carlo simulation setting as that of the comparison for different CRL methods, and we also use the same error metrics as in \eqref{eq:error_statistical_criterion}. The three methods are abbreviated as fCRL(MLVC), fCRL(MVC), and fCRL(MGC) for the CRL methods using the LV kernel, the Versoria kernel, and the Gaussian kernel, respectively. The violin statistical plots are given in Fig. \ref{fig:violin_kernel}, where Fig. \ref{fig:violin_kernel_angle} and Fig. \ref{fig:violin_kernel_dis} summarize the $\text{TR-ER}_{\psi}$ ($\text{SS-ER}_{\psi}$) and $\text{TR-ER}_{\p}$ ($\text{SS-ER}_{\p}$) data, respectively. Likewise, the bootstrapping plots are given in \ref{fig:boot_kernel}. According to Fig. \ref{fig:violin_kernel} and Fig. \ref{fig:boot_kernel}, there is no much difference between the CRL methods using different kernels, especially for steady-state performance.
\begin{figure}
    \centering
  \subfloat[Heading Error]{%
       \includegraphics[width=\linewidth]{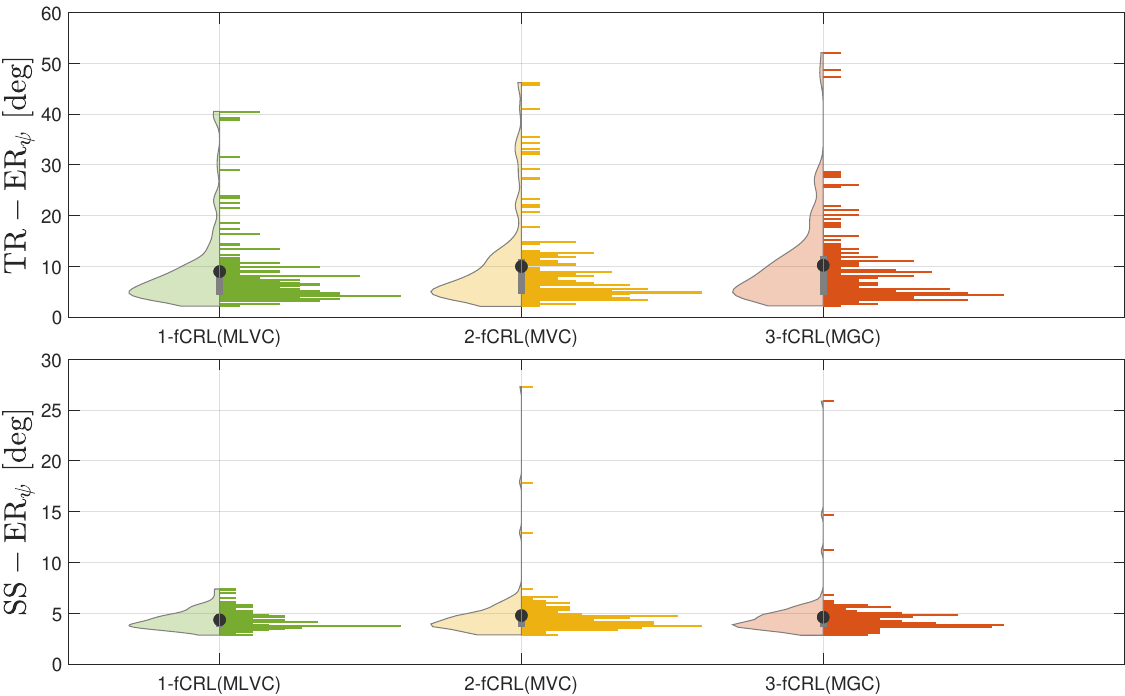}
       \label{fig:violin_kernel_angle}}
    \\
  \subfloat[Position Error]{%
        \includegraphics[width=\linewidth]{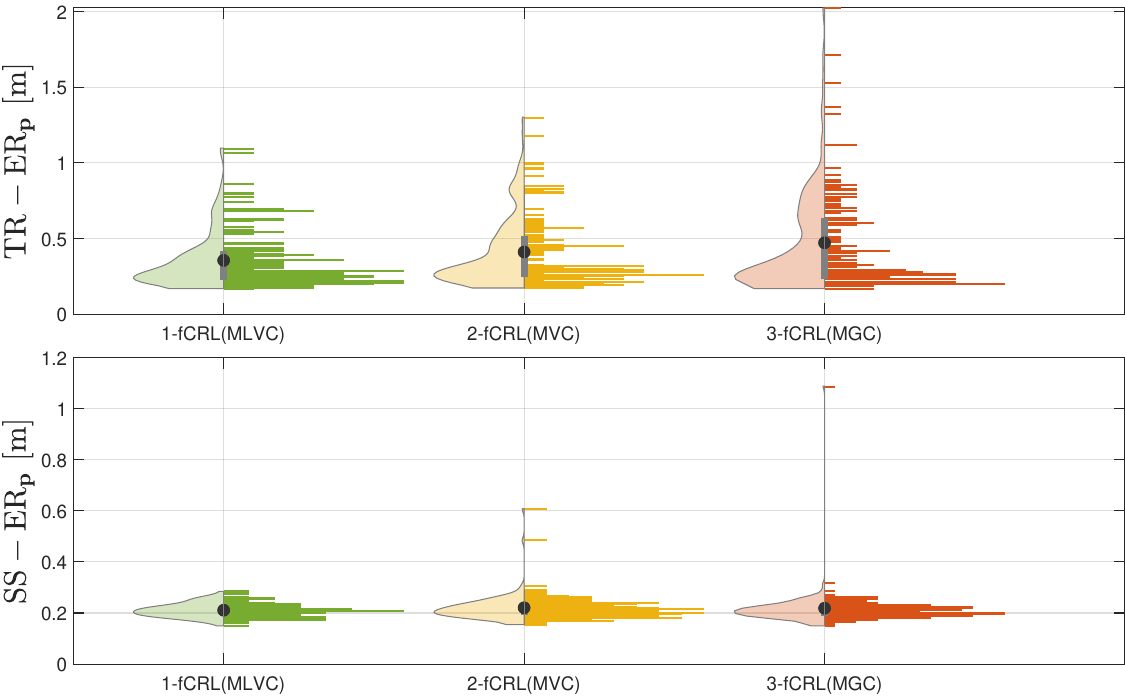}
        \label{fig:violin_kernel_dis}}
  \caption{Violin Plots of the Monte Carlo simulation for the 3 CRL methods with the filters in \textbf{smart} mode, (a) Averaged Heading Error (b) Averaged Position Error. For each subfigure (a) and (b), from left to right: 1) fCRL(MLVC) 2) fCRL(MVC) 3) fCRL(MGC). \chang{For each violin plot, the averaged error metric concerning the trials is marked as a black dot, the left part is the approximated PDF of the data with the two boundaries of the shaded area representing the $25\%$ and $75\%$ quantile, and the right part is the histogram.}}
  \label{fig:violin_kernel} 
\end{figure}
\begin{figure}
    \centering
    \includegraphics[width=0.45\textwidth]{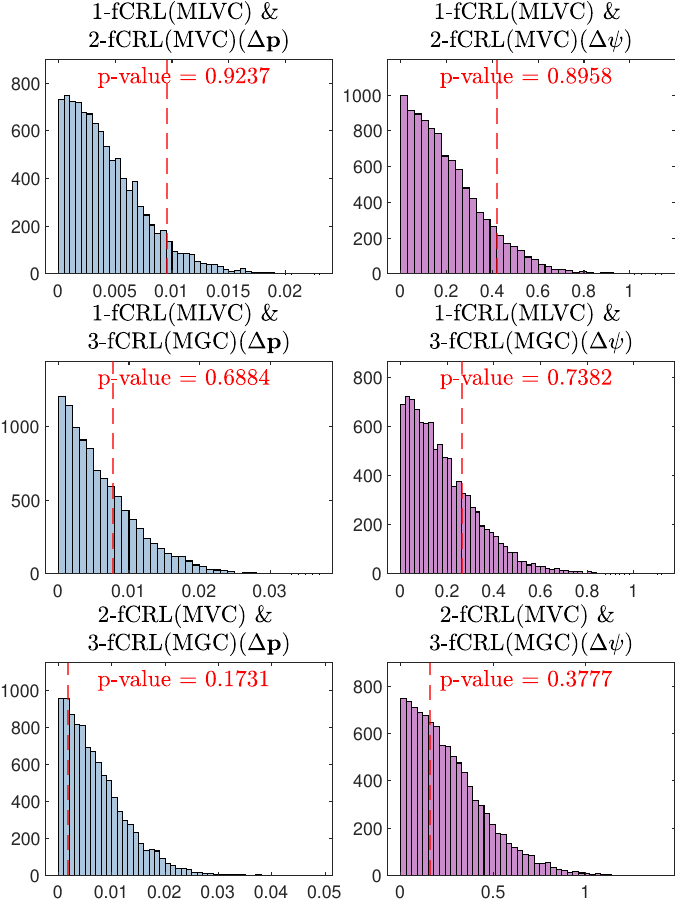}
    \caption{Bootstrapping statistical analysis for steady-state performance of the methods fCRL(MLVC), fCRL(MVC), and fCRL(MGC).}
    \label{fig:boot_kernel}
\end{figure}
Another important aspect that requires evaluation in kernel-based EKF is the number of fixed-point iterations. For each of the three methods, there are $3\times 10^{3}$ \guido{time stamps} for each trial, thus there are $3\times 10^{3} \times 120 = 3.6 \times 10^{5}$ \guido{data for the number of fixed-point iterations} in total for the $120$ trials. The histogram of the $3.6 \times 10^{5}$ numbers for each of the three methods \chang{is} presented in Fig. \ref{fig:histo_counting}, where the average number of the fixed-point iteration is also given. From the three histograms in Fig. \ref{fig:histo_counting}, it is clear that the LV kernel outperforms the Versoria kernel and the Gaussian kernel for having fewer number of fixed-point iterations, which is highly desired for running the estimation algorithm online with limited computational resources.
\begin{figure}
    \centering
    \includegraphics[width=0.45\textwidth]{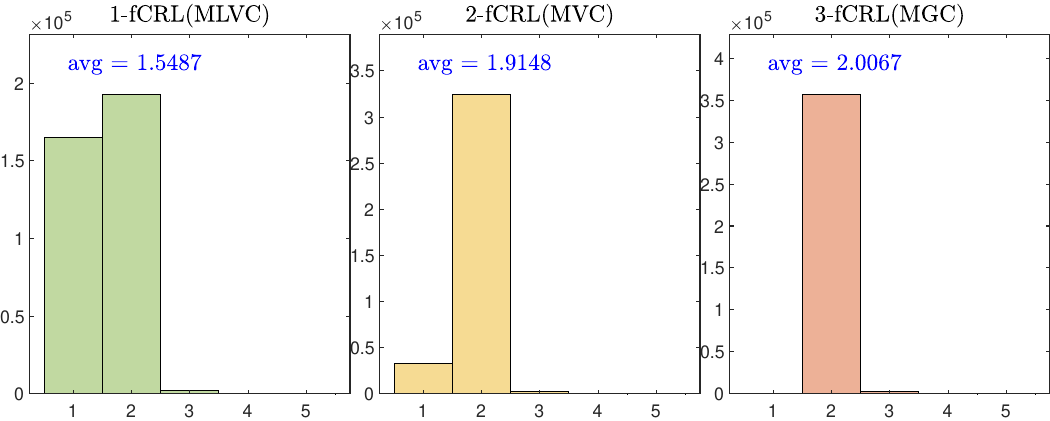}
    \caption{Histogram of the number of fixed-point iterations. From left to right: 1) fCRL(MLVC) 2) fCRL(MVC) 3) fCRL(MGC).}
    \label{fig:histo_counting}
\end{figure}
}
\input{content_file/5.3.1-extreme_kernel.tex}

%% file: content_file/5.3.1-extreme_kernel.tex
\chang{
\begin{figure*}
    \centering
    \includegraphics[width=\textwidth]{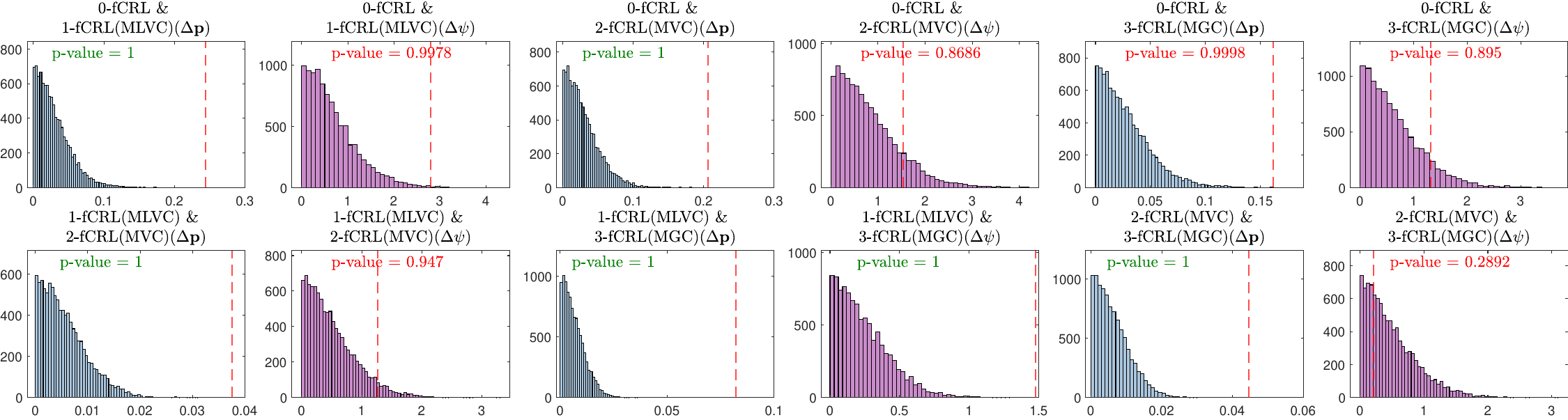}
    \caption{Bootstrapping statistical analysis for steady-state performance of the methods fCRL, fCRL(MLVC), fCRL(MVC), and fCRL(MGC). For fCRL(MLVC), fCRL(MVC), and fCRL(MGC), the number of allowed fixed-point iterations is restricted to one.}
    \label{fig:boot_exk}
\end{figure*}
\begin{figure}
    \centering
  \subfloat[Heading Error]{%
       \includegraphics[width=\linewidth]{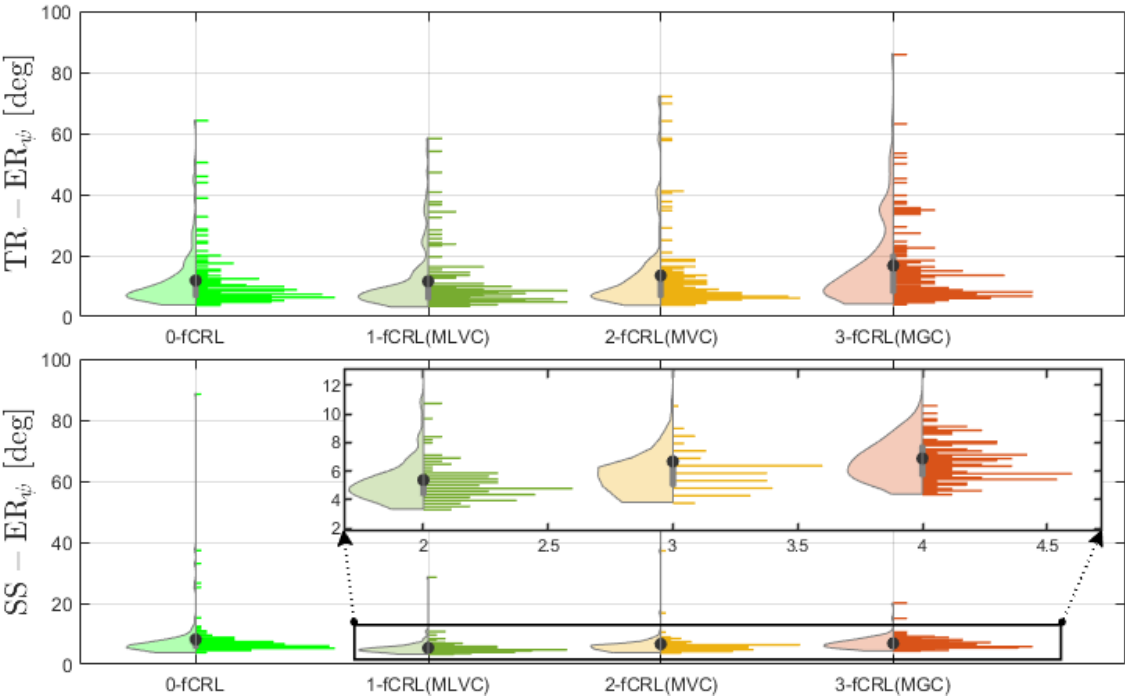}
       \label{fig:violin_exk_angle}}
    \\
  \subfloat[Position Error]{%
        \includegraphics[width=\linewidth]{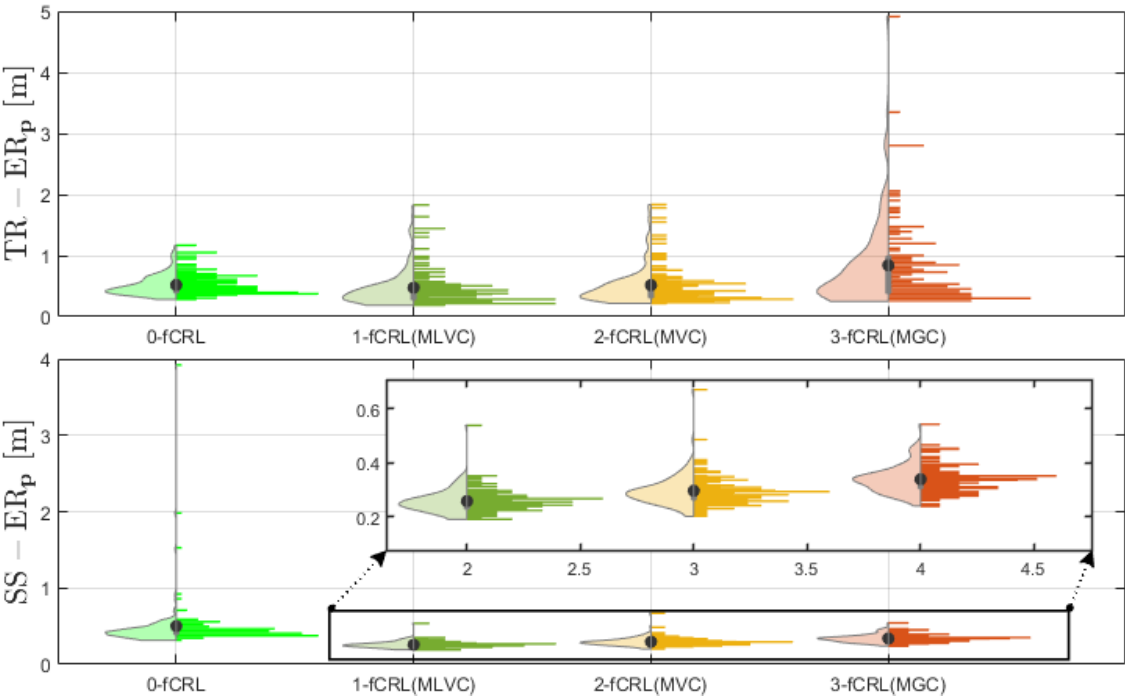}
        \label{fig:violin_exk_dis}}
  \caption{Violin Plots of the Monte Carlo simulation for the kernel-induced EKF-based fCRL localization methods with only one fixed-point iteration (a) Averaged Heading Error (b) Averaged Position Error. For each subfigure (a) and (b), from left to right: 1) fCRL, 2) fCRL(MLVC) 3) fCRL(MVC) 4) fCRL(MGC). \chang{For each violin plot, the averaged error metric concerning the trials is marked as a black dot, the left part is the approximated PDF of the data with the two boundaries of the shaded area representing the $25\%$ and $75\%$ quantile, and the right part is the histogram.}}
  \label{fig:violin_exk} 
\end{figure}

A possible compromise is to restrict the fixed-point iteration to run \textit{only once} at each time instant $k$, which would further reduce the computational complexity and is more suitable for running the estimation algorithm online. To this end, another round of Monte Carlo simulation was carried out where the methods fCRL(MLVC), fCRL(MVC), fCRL(MGC) are compared and the number of fixed-point iterations for all three methods is uniformly set to one. In addition, the results of the fCRL method (i.e., fCRL with the EKF), serving as the baseline method, are also included for comparison purposes. The corresponding violin plots are shown in Fig. \ref{fig:violin_exk}, and the accompanying bootstrapping results are given in Fig. \ref{fig:boot_exk}, from which it is clear that the fCRL(MLVC) outperforms the other methods in terms of the position estimation for both the transient and steady-state intervals.
}

%% file: content_file/6-conclusion.tex
\section{Conclusion \& Future Work}
\label{sec:6-conclusion}
In this work, a new CRL scheme is proposed for the localization task of an agent which aims to localize all its neighbors in its body frame. The new CRL scheme can account for the correlation induced by the same velocity input and benefit from indirect distance measurements between the neighbors. Observability analysis using the augmented Lie derivative is carried out on the CRL model, which shows that bringing additional indirect measurements expands the observable subspace. To handle the heavy-tailed UWB noise, the kernel-induced Kalman filter with a novel-designed LV kernel is applied to the state estimation problem. Sufficient conditions for the convergence of the fixed-point iteration in the filtering algorithm are derived. The advantages of using the proposed CRL method in combination with the kernel-induced EKF are demonstrated through a comparative study. Simulation results show that the proposed localization method outperforms its other variations in terms of improved estimation accuracy, robustness against measurement outliers, and insensitivity to measurement covariance matrix initialization. Moreover, the proposed filtering with the LV kernel also outperforms those using other kernels in extreme cases when the fixed-point iteration degenerates to the one-step kernel-induced measurement update.

%% file: reference.tex
{
\tiny
\bibliographystyle{IEEEtran}
\bibliography{bibliography.bib}
}

%% file: bare_jrnl_new_sample4.bbl
\begin{thebibliography}{10}
\providecommand{\url}[1]{#1}
\csname url@samestyle\endcsname
\providecommand{\newblock}{\relax}
\providecommand{\bibinfo}[2]{#2}
\providecommand{\BIBentrySTDinterwordspacing}{\spaceskip=0pt\relax}
\providecommand{\BIBentryALTinterwordstretchfactor}{4}
\providecommand{\BIBentryALTinterwordspacing}{\spaceskip=\fontdimen2\font plus
\BIBentryALTinterwordstretchfactor\fontdimen3\font minus
  \fontdimen4\font\relax}
\providecommand{\BIBforeignlanguage}[2]{{%
\expandafter\ifx\csname l@#1\endcsname\relax
\typeout{** WARNING: IEEEtran.bst: No hyphenation pattern has been}%
\typeout{** loaded for the language `#1'. Using the pattern for}%
\typeout{** the default language instead.}%
\else
\language=\csname l@#1\endcsname
\fi
#2}}
\providecommand{\BIBdecl}{\relax}
\BIBdecl

\bibitem{scaramuzza2014vision}
D.~Scaramuzza, M.~C. Achtelik, L.~Doitsidis, F.~Friedrich, E.~Kosmatopoulos,
  A.~Martinelli, M.~W. Achtelik, M.~Chli, S.~Chatzichristofis, L.~Kneip
  \emph{et~al.}, ``Vision-controlled micro flying robots: from system design to
  autonomous navigation and mapping in gps-denied environments,'' \emph{IEEE
  Robotics \& Automation Magazine}, vol.~21, no.~3, pp. 26--40, 2014.

\bibitem{bahr2009consistent}
A.~Bahr, M.~R. Walter, and J.~J. Leonard, ``Consistent cooperative
  localization,'' in \emph{2009 IEEE International Conference on Robotics and
  Automation}.\hskip 1em plus 0.5em minus 0.4em\relax IEEE, 2009, pp.
  3415--3422.

\bibitem{augugliaro2014flight}
F.~Augugliaro, S.~Lupashin, M.~Hamer, C.~Male, M.~Hehn, M.~W. Mueller, J.~S.
  Willmann, F.~Gramazio, M.~Kohler, and R.~D'Andrea, ``The flight assembled
  architecture installation: Cooperative construction with flying machines,''
  \emph{IEEE Control Systems Magazine}, vol.~34, no.~4, pp. 46--64, 2014.

\bibitem{menouar2017uav}
H.~Menouar, I.~Guvenc, K.~Akkaya, A.~S. Uluagac, A.~Kadri, and A.~Tuncer,
  ``Uav-enabled intelligent transportation systems for the smart city:
  Applications and challenges,'' \emph{IEEE Communications Magazine}, vol.~55,
  no.~3, pp. 22--28, 2017.

\bibitem{duisterhof2021sniffy}
B.~P. Duisterhof, S.~Li, J.~Burgu{\'e}s, V.~J. Reddi, and G.~C. de~Croon,
  ``Sniffy bug: A fully autonomous swarm of gas-seeking nano quadcopters in
  cluttered environments,'' in \emph{2021 IEEE/RSJ International Conference on
  Intelligent Robots and Systems (IROS)}.\hskip 1em plus 0.5em minus
  0.4em\relax IEEE, 2021, pp. 9099--9106.

\bibitem{coppola2018board}
M.~Coppola, K.~N. McGuire, K.~Y. Scheper, and G.~C. de~Croon, ``On-board
  communication-based relative localization for collision avoidance in micro
  air vehicle teams,'' \emph{Autonomous robots}, vol.~42, no.~8, pp.
  1787--1805, 2018.

\bibitem{guo2019ultra}
K.~Guo, X.~Li, and L.~Xie, ``Ultra-wideband and odometry-based cooperative
  relative localization with application to multi-uav formation control,''
  \emph{IEEE transactions on cybernetics}, vol.~50, no.~6, pp. 2590--2603,
  2019.

\bibitem{dong2016time}
X.~Dong, Y.~Zhou, Z.~Ren, and Y.~Zhong, ``Time-varying formation tracking for
  second-order multi-agent systems subjected to switching topologies with
  application to quadrotor formation flying,'' \emph{IEEE Transactions on
  Industrial Electronics}, vol.~64, no.~6, pp. 5014--5024, 2016.

\bibitem{kang2016distance}
S.-M. Kang, M.-C. Park, and H.-S. Ahn, ``Distance-based cycle-free persistent
  formation: Global convergence and experimental test with a group of
  quadcopters,'' \emph{IEEE Transactions on Industrial Electronics}, vol.~64,
  no.~1, pp. 380--389, 2016.

\bibitem{dudzik2020application}
S.~Dudzik, ``Application of the motion capture system to estimate the accuracy
  of a wheeled mobile robot localization,'' \emph{Energies}, vol.~13, no.~23,
  p. 6437, 2020.

\bibitem{fang2020graph}
X.~Fang, C.~Wang, T.-M. Nguyen, and L.~Xie, ``Graph optimization approach to
  range-based localization,'' \emph{IEEE Transactions on Systems, Man, and
  Cybernetics: Systems}, vol.~51, no.~11, pp. 6830--6841, 2020.

\bibitem{ruan2021cooperative}
L.~Ruan, G.~Li, W.~Dai, S.~Tian, G.~Fan, J.~Wang, and X.~Dai, ``Cooperative
  relative localization for uav swarm in gnss-denied environment: A coalition
  formation game approach,'' \emph{IEEE Internet of Things Journal}, 2021.

\bibitem{nguyen2016ultra}
T.~M. Nguyen, A.~H. Zaini, K.~Guo, and L.~Xie, ``An ultra-wideband-based
  multi-uav localization system in gps-denied environments,'' in \emph{2016
  International Micro Air Vehicles Conference}, vol.~6, 2016, pp. 1--15.

\bibitem{chudoba2014localization}
J.~Chudoba, M.~Saska, T.~B{\'a}{\v{c}}a, and L.~P{\v{r}}eu{\v{c}}il,
  ``Localization and stabilization of micro aerial vehicles based on visual
  features tracking,'' in \emph{2014 International Conference on Unmanned
  Aircraft Systems (ICUAS)}.\hskip 1em plus 0.5em minus 0.4em\relax IEEE, 2014,
  pp. 611--616.

\bibitem{jurevivcius2019robust}
R.~Jurevi{\v{c}}ius, V.~Marcinkevi{\v{c}}ius, and J.~{\v{S}}eibokas, ``Robust
  gnss-denied localization for uav using particle filter and visual odometry,''
  \emph{Machine Vision and Applications}, vol.~30, no.~7, pp. 1181--1190, 2019.

\bibitem{guler2020peer}
S.~G{\"u}ler, M.~Abdelkader, and J.~S. Shamma, ``Peer-to-peer relative
  localization of aerial robots with ultrawideband sensors,'' \emph{IEEE
  Transactions on Control Systems Technology}, vol.~29, no.~5, pp. 1981--1996,
  2020.

\bibitem{li2020autonomous}
S.~Li, M.~Coppola, C.~De~Wagter, and G.~C. de~Croon, ``An autonomous swarm of
  micro flying robots with range-based relative localization,'' \emph{arXiv
  preprint arXiv:2003.05853}, 2020.

\bibitem{lee2007comparative}
J.-S. Lee, Y.-W. Su, and C.-C. Shen, ``A comparative study of wireless
  protocols: Bluetooth, uwb, zigbee, and wi-fi,'' in \emph{IECON 2007-33rd
  Annual Conference of the IEEE Industrial Electronics Society}.\hskip 1em plus
  0.5em minus 0.4em\relax Ieee, 2007, pp. 46--51.

\bibitem{shalaby2021relative}
M.~Shalaby, C.~C. Cossette, J.~R. Forbes, and J.~Le~Ny, ``Relative position
  estimation in multi-agent systems using attitude-coupled range
  measurements,'' \emph{IEEE Robotics and Automation Letters}, vol.~6, no.~3,
  pp. 4955--4961, 2021.

\bibitem{liu2020distributed}
Y.~Liu, Y.~Wang, J.~Wang, and Y.~Shen, ``Distributed 3d relative localization
  of uavs,'' \emph{IEEE Transactions on Vehicular Technology}, vol.~69, no.~10,
  pp. 11\,756--11\,770, 2020.

\bibitem{trawny20073d}
N.~Trawny, X.~S. Zhou, K.~X. Zhou, and S.~I. Roumeliotis, ``3d relative pose
  estimation from distance-only measurements,'' in \emph{2007 IEEE/RSJ
  International Conference on Intelligent Robots and Systems}.\hskip 1em plus
  0.5em minus 0.4em\relax IEEE, 2007, pp. 1071--1078.

\bibitem{jiang20193}
B.~Jiang, B.~D. Anderson, and H.~Hmam, ``3-d relative localization of mobile
  systems using distance-only measurements via semidefinite optimization,''
  \emph{IEEE Transactions on Aerospace and Electronic Systems}, vol.~56, no.~3,
  pp. 1903--1916, 2019.

\bibitem{chen2017maximum}
B.~Chen, X.~Liu, H.~Zhao, and J.~C. Principe, ``Maximum correntropy kalman
  filter,'' \emph{Automatica}, vol.~76, pp. 70--77, 2017.

\bibitem{pfeiffer2021computationally}
S.~Pfeiffer, C.~De~Wagter, and G.~C. De~Croon, ``A computationally efficient
  moving horizon estimator for ultra-wideband localization on small
  quadrotors,'' \emph{IEEE Robotics and Automation Letters}, vol.~6, no.~4, pp.
  6725--6732, 2021.

\bibitem{isidori1985nonlinear}
A.~Isidori, \emph{Nonlinear control systems, 3rd ed.}\hskip 1em plus 0.5em
  minus 0.4em\relax London, Springer Verlag, 1995.

\bibitem{martinelli2022extension}
A.~Martinelli, ``Extension of the observability rank condition to time-varying
  nonlinear systems,'' \emph{IEEE Transactions on Automatic Control}, 2022.

\bibitem{martinelli2005observability}
A.~Martinelli and R.~Siegwart, ``Observability analysis for mobile robot
  localization,'' in \emph{2005 IEEE/RSJ International Conference on
  Intelligent Robots and Systems}.\hskip 1em plus 0.5em minus 0.4em\relax IEEE,
  2005, pp. 1471--1476.

\bibitem{reif1999stochastic}
K.~Reif, S.~Gunther, E.~Yaz, and R.~Unbehauen, ``Stochastic stability of the
  discrete-time extended kalman filter,'' \emph{IEEE Transactions on Automatic
  control}, vol.~44, no.~4, pp. 714--728, 1999.

\bibitem{van2021control}
J.~H. van Schuppen, \emph{Control and System Theory of Discrete-Time Stochastic
  Systems}.\hskip 1em plus 0.5em minus 0.4em\relax Springer, 2021.

\bibitem{principe2010information}
J.~C. Principe, \emph{Information theoretic learning: Renyi's entropy and
  kernel perspectives}.\hskip 1em plus 0.5em minus 0.4em\relax Springer Science
  \& Business Media, 2010.

\bibitem{huang2017maximum}
F.~Huang, J.~Zhang, and S.~Zhang, ``Maximum versoria criterion-based robust
  adaptive filtering algorithm,'' \emph{IEEE Transactions on Circuits and
  Systems II: Express Briefs}, vol.~64, no.~10, pp. 1252--1256, 2017.

\bibitem{agarwal2001fixed}
R.~P. Agarwal, M.~Meehan, and D.~O'regan, \emph{Fixed point theory and
  applications}.\hskip 1em plus 0.5em minus 0.4em\relax Cambridge university
  press, 2001, vol. 141.

\end{thebibliography}
